%% file: neurips_2026.tex
\documentclass{article}

\usepackage[numbers,square,compress]{natbib}
\usepackage[preprint]{neurips_2026}

\usepackage[T1]{fontenc}    
\usepackage{hyperref}       
\usepackage{url}            
\usepackage{booktabs}       
\usepackage{microtype}      
\usepackage{xcolor}         

\usepackage{marvosym}

\usepackage{makecell}
\usepackage{graphicx}
\usepackage{float}
\usepackage{adjustbox}
\usepackage{amsmath}
\usepackage{amssymb}
\usepackage{wrapfig}
\usepackage{enumitem}
\usepackage{mathtools}
\usepackage{array}
\usepackage{colortbl}
\usepackage[most]{tcolorbox}
\usepackage[toc,page]{appendix}

\definecolor{oai-green}{HTML}{51DA4C}
\definecolor{oai-magenta}{HTML}{FF45FF}
\definecolor{oai-gray-200}{HTML}{F5F5F5}

\newcommand{\infobox}[1]{
    \begin{tcolorbox}[
        colback=white!90!gray,
        colframe=teal!60!black,
        arc=5pt, boxsep=5pt,
        left=5pt, right=10pt, top=2pt, bottom=3pt,
        boxrule=0.8pt,
        drop shadow=gray!50!white,
        enhanced jigsaw
    ]
         \textit{#1}
    \end{tcolorbox}
}

\title{$M^3Eval$: Multi-Modal Memory Evaluation \\ through Cognitively-Grounded Video Tasks}

\author{%
Jie Huang$^{\star,1,2}$,
Ruixun Liu$^{\star,1,2}$,
Sirui Sun$^{3}$,
Xinyi Yang$^{4,5,2}$, \\
\textbf{Yin Li$^{6}$,
Yixin Zhu$^{5,4,2}$,
Yiwu Zhong$^{1,2,\textrm{\Letter}}$} \\
$^{\star}$ equal contributors \quad\quad $\textrm{\Letter}$ corresponding author \\
$^{1}$ School of Intelligence Science and Technology, Peking University \\
$^{2}$ State Key Laboratory of General Artificial Intelligence, Peking University \\
$^{3}$ Yuanpei College, Peking University \\
$^{4}$ Institute for Artificial Intelligence, Peking University \\
$^{5}$ School of Psychological and Cognitive Sciences, Peking University \\
$^{6}$ University of Wisconsin-Madison
}

\begin{document}

\maketitle

\input{sections/00_abstract}
\input{sections/01_introduction}
\input{sections/02_REVISED}
\input{sections/03_experimental_setup}
\input{sections/04_experiments}

\input{sections/06_conclusion}

\bibliographystyle{plainnat}
\bibliography{references}

\clearpage
\appendix
\input{suppl_content_new}
\input{Appendix/qa_construction}

\end{document}

%% file: sections/00_abstract.tex
\begin{abstract}

As multi-modal models advance towards long-form video understanding, memory emerges as a critical capability. Despite substantial efforts in developing video datasets and benchmarks, existing works primarily focus on perception and reasoning, without systematically evaluating memory: what models retain, how faithfully information is preserved, and how robust memory remains under interference.
To address this gap, we introduce $M^3Eval$, the first comprehensive evaluation framework and benchmark for probing different memory dimensions in multi-modal models.
Grounded in cognitive psychology, our design features carefully constructed tasks that isolate key aspects of memory. Leveraging $M^3 Eval$, we conduct extensive experiments across representative multi-modal models, revealing consistent weaknesses and distinctive behaviors.
We find that models struggle to maintain disentangled representations when processing parallel video streams, exhibit interference patterns differing substantially from those observed in human memory, ground memory sources more reliably in the spatial domain than the temporal domain, and demonstrate limited symbolic memory.
Collectively, our benchmark provides a valuable resource for future research, while our findings highlight memory as a fundamental yet underexplored capability and offer insights for designing more effective memory mechanisms in multi-modal models. 
Our code and dataset are available at \url{https://pku-value-lab.github.io/m3eval-homepage}.

\end{abstract}

%% file: sections/01_introduction.tex
\section{Introduction}

Multi-modal models~\cite{bai2025qwen3,internvl35,qwen35,gpt54,gemini3pro} are rapidly advancing towards long-form video understanding, driven in part by expanding context windows. However, increasing context alone does not guarantee effective memory. A core challenge lies in the memory mechanism itself~\cite{hu2025memory, jia2026ai, liang2025ai,zhang2025survey}, the ability to encode, store, retrieve, and synthesize information over long temporal horizons spanning both video and text. Such memory is critical for retaining information across long video streams and multi-turn interactions, and for enabling downstream reasoning that depends on this information~\cite{m3agent,videolucy,fan2024videoagent,song2024moviechat,song2025moviechat+}. Despite growing interest in this capability, there is currently no dedicated evaluation protocol or benchmark for systematically probing memory in multi-modal models. As a result, their memory capabilities remain poorly measured and not well understood.

A large body of multi-modal datasets and benchmarks has been developed for video understanding~\cite{fu2025video, li2024mvbench}. These benchmarks primarily focus on visual perception and reasoning. While some tasks implicitly involve memory, for example, long video understanding~\cite{zhou2025mlvu, wang2025lvbench, chandrasegaran2024hourvideo, zhou2025x, yang2025egolife} or video reasoning~\cite{cheng2025video,yang2025svbench}, they are not designed to isolate memory mechanisms. Consequently, they provide only a partial and indirect assessment of memory. In particular, existing benchmarks rarely disentangle different aspects of memory, such as \textit{capacity} (how much information can be retained), \textit{fidelity} (how accurately stored information is preserved), and \textit{robustness} (how well representations withstand interference from similar or distracting inputs). 

\begin{figure}[t]
    \centering
    \includegraphics[width=1\textwidth]{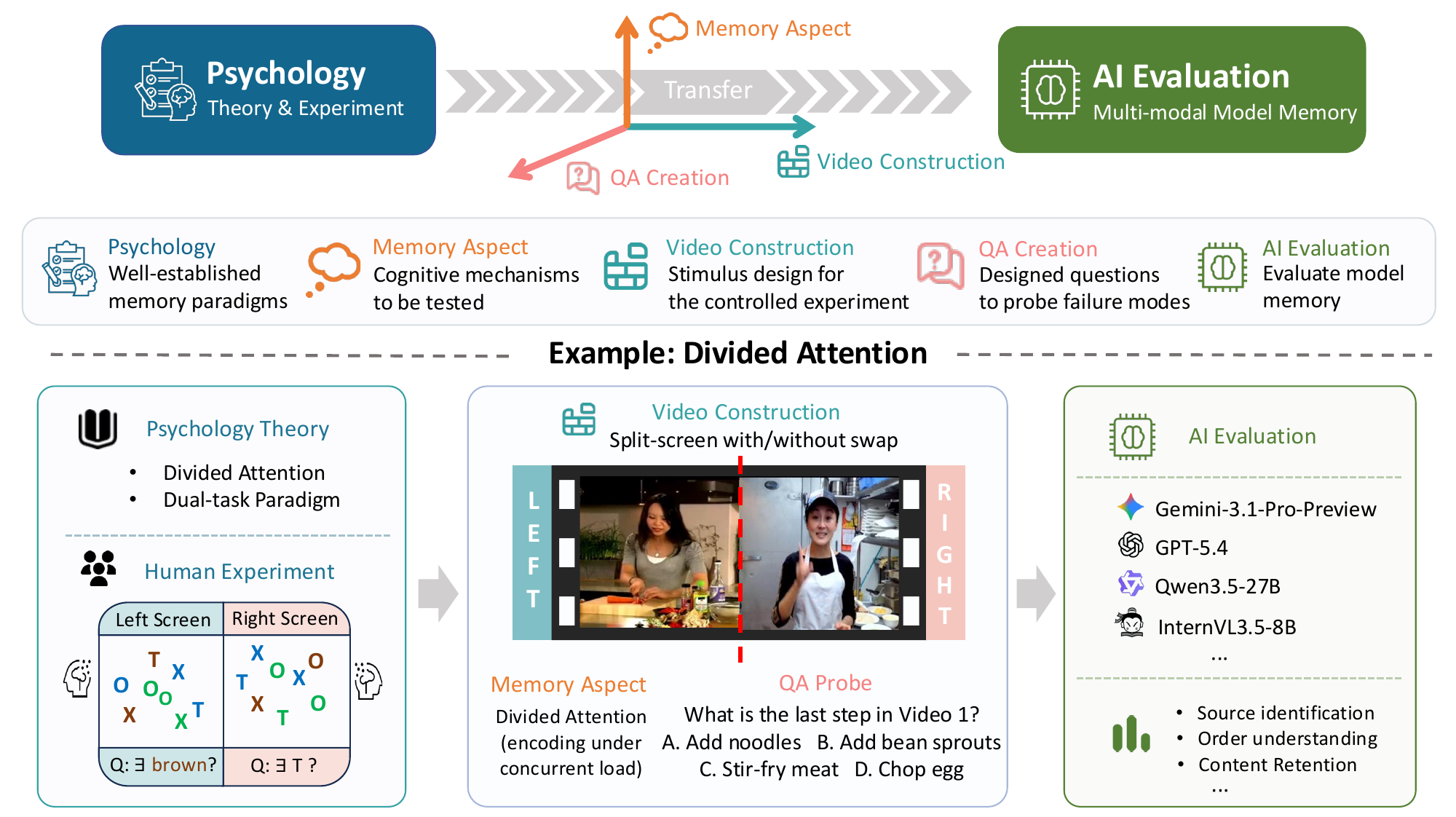}
    \vspace{-7mm}
    \caption{\textbf{$M^3Eval$}, our principled framework and benchmark for evaluating memory capabilities of multi-modal models. We present an example task of \textit{divided attention}. Grounded in psychological theory, we construct split-screen video scenarios, design memory questions, and analyze multiple models in terms of source identification, order understanding, and content retention.}
    \label{fig:intro}
    \vspace{-6mm}
\end{figure}

To address this gap, we introduce $M^3Eval$, a principled evaluation framework and benchmark for probing memory capabilities in multi-modal models.
As illustrated in Fig.~\ref{fig:intro}, our design is inspired by the controlled experimental paradigms in cognitive psychology~\cite{kahana2024oxford,schwieter2022cambridge}, where memory is studied through carefully constructed stimuli that isolate specific mechanisms. We adapt these principles to video domain by constructing video-based QA tasks that probe memory under controlled yet realistic conditions. Our benchmark characterizes memory along four key dimensions: \textit{(1)} the ability to retain information from concurrent inputs~\cite{baddeley1998working,craik1996effects,treisman1980feature,treisman1982illusory}; \textit{(2)} robustness to interference from similar content~\cite{mcgeoch1932forgetting,underwood1957interference,zaromb2006temporal,Robinson1920}; \textit{(3)} the ability to integrate interleaved events into coherent representations~\cite{mandler1977remembrance,mandler1978code,johnson1993source,schacter1984retrieval}; and \textit{(4)} the ability to track abstract attributes across video segments~\cite{kirchner1958age,owen2005nback}. While grounded in cognitive theory, these dimensions also arise in real-world video understanding, such as scene analysis, object tracking, and long context reasoning.

Leveraging $M^3Eval$, we conduct an extensive evaluation on both open-source and proprietary multi-modal models. Our results reveal several notable and, in some cases, unexpected findings. 
\textit{First}, when processing parallel video streams, the models fail to maintain independent representations for each stream; we hypothesize that such failure stems from attention confusion across concurrent visual inputs. 
\textit{Second}, humans exhibit notably stronger retroactive interference than proactive interference, whereas multi-modal models demonstrate comparable interference levels. This contrast indicates a fundamental difference in memory mechanisms between humans and models. Surprisingly, repeating interfering video segments can even improve model understanding about the target video segments.
\textit{Third}, model memory is less capable than human memory when organizing temporally interleaved information. Further analysis reveals that memory source grounding along temporal dimension is consistently weaker than spatial dimension. 
\textit{Finally}, the models exhibit far weaker symbolic memory than humans when required to abstract multi-modal information into symbolic attributes and distinguish their relations. We further find that the models struggle to filter out irrelevant information from memory.

\newpage
\smallskip
\textbf{Our contributions} are summarized as follows.
\begin{itemize}
    \item We introduce $M^3Eval$, \textbf{the first benchmark} for systematically evaluating different dimensions of memory capabilities of multi-modal models with video tasks.
    \item \textbf{Our key innovation} lies in a cognitively-grounded evaluation design that isolates memory mechanisms through orchestrated video tasks.
    \item We provide \textbf{a systematic evaluation} across diverse models, offering new insights into the limitations of current multi-modal memory and informing the design of future systems.
\end{itemize}

%% file: sections/02_REVISED.tex
\section{Related Work}

\textbf{Memory Evaluation in LLMs and Agents.}
The evaluation of memory capabilities has been recently studied for LLMs and LLM-based agents~\cite{hu2025memory, jia2026ai, liang2025ai,zhang2025survey}. Early benchmarks relied on synthetic needle-in-a-haystack tasks~\cite{song2024counting,hsieh2024ruler,kuratov2024babilong} or long-range dialogues~\cite{maharana2024evaluating,jia2025evaluating} to assess retention within a fixed context. Dynamic benchmarks~\cite{tan2025membench,wu2024longmemeval} further required incremental memory updates across turns. Wei et al.~\cite{wei2025evo} and Zhang et al.~\cite{zheng2025lifelongagentbench} introduced self-evolution settings to examine whether models can distill strategies from past experience.
While the above efforts focus primarily on text, Mem-Gallery~\cite{bei2026mem} extends memory evaluation to the multimodal setting with multi-session dialogues grounded in both text and images.
Inspired by cognitive psychology, recent studies~\cite{gong2024working,zhang2024working} adopted the N-Back task~\cite{kirchner1958age} to assess working memory capacity.
However, none of these works has explored memory evaluation for video tasks.

\smallskip
\textbf{Evaluation for Video Understanding.}
Memory is an essential yet underexplored component for video understanding. 
Numerous benchmarks evaluate general video understanding~\cite{fu2025video, li2024mvbench}, long-form video tasks~\cite{zhou2025mlvu, wang2025lvbench, chandrasegaran2024hourvideo, zhou2025x, yang2025egolife}, streaming evaluation~\cite{niu2025ovo, lin2026streamingbench, xiong2025streaming}, and cross-video understanding~\cite{zhu2025cvbench, li2026crossvid}. However, these benchmarks often conflate memory with visual perception and reasoning, treating memory as an implicit component rather than measuring it explicitly. Another line adopts synthetic needle-in-a-haystack settings~\cite{zhao2024needle, xia2025video, hu2025nemo, yang2025cambrian, li2025two}, inserting target segments into distractor footage to test retrieval over extended contexts. Yet these approaches rely on simple probe designs, making it difficult to assess different dimensions of memory.
A recent effort~\cite{m3agent} probes memory through reasoning tasks, yet does not directly and systematically evaluate memory across multiple dimensions.
Unlike these works, our benchmark leverages existing video datasets and explicitly probes key dimensions of memory through cognitively-grounded evaluation paradigms.

\smallskip
\textbf{Memory Investigation in Cognitive Psychology.}
Cognitive psychology decomposes memory into distinct, measurable processes. Our evaluation framework builds on four such processes:
\textit{(1)~Divided Attention.} Divided attention during encoding degrades retention and induces illusory conjunctions~\cite{craik1996effects,treisman1980feature,treisman1982illusory}, as the cognitive resources for encoding are limited~\cite{Kahneman1973-KAHAAE,baddeley1998working}.
\textit{(2)~Memory Interference.} Forgetting arises from competition among similar memory traces rather than simple decay. Such competition manifests as proactive or retroactive interference~\cite{mcgeoch1932forgetting,underwood1957interference,zaromb2006temporal, Robinson1920}.
\textit{(3)~Memory Organization.} Recall relies on implicit story schemata~\cite{mandler1977remembrance}. When processing interleaved storylines, individuals default to the underlying event structure~\cite{mandler1978code}.
\textit{(4)~N-Back and Symbolic Representation.} The N-Back task~\cite{kirchner1958age,owen2005nback} is widely used to isolate memory capability and reflects the view that memory operates over abstract representations~\cite{anderson2014human,pylyshyn1973mind}.

%% file: sections/03_experimental_setup.tex
\section{Memory Evaluation}
\vspace{-2mm}
\label{sec:eval_design}

\begin{figure}[H]
    \centering    \includegraphics[width=0.9\textwidth]{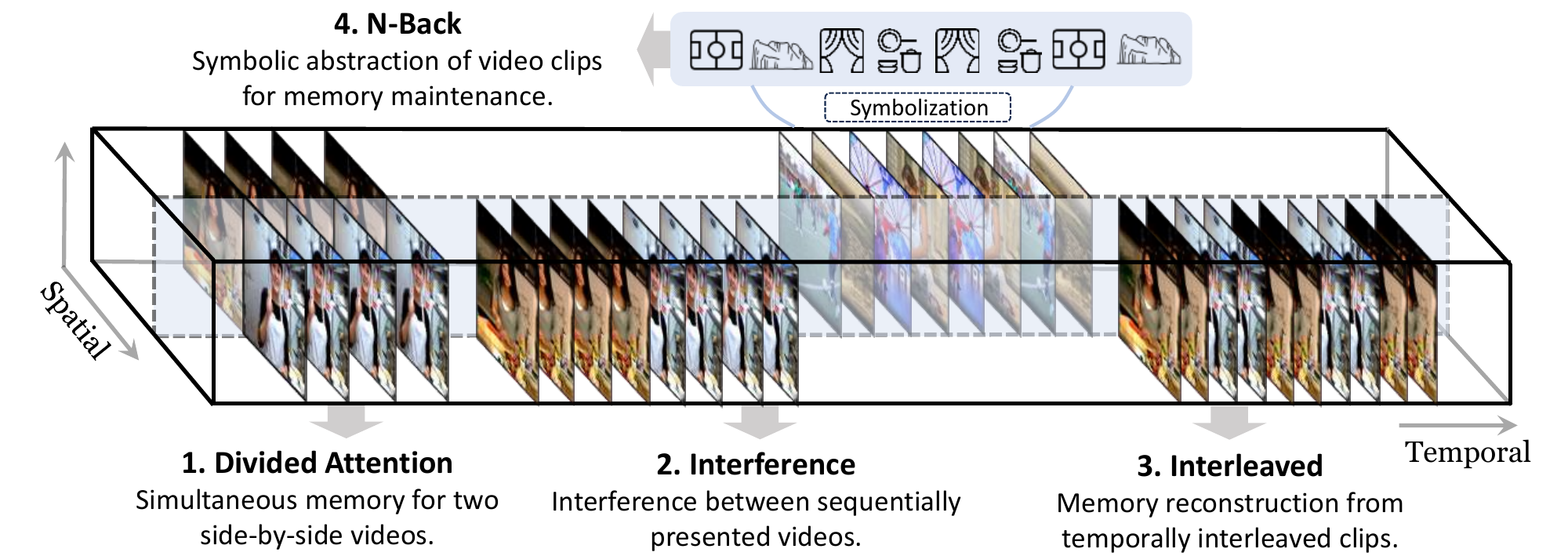}
    \vspace{-3mm}
    \caption{\textbf{Overview} of the unified and coherent framework for our four evaluation paradigms.} 
    \label{fig:methodfinal}
\end{figure}

As shown in Fig.~\ref{fig:methodfinal}, our evaluation consists of four paradigms unified under a coherent framework. Along the spatial dimension, \textit{Divided Attention} evaluates the encoding under concurrent visual inputs. As for temporal dimension, \textit{Memory Interference} tests robustness to the distraction from sequential similar content, while \textit{Interleaved Events} examines temporal reorganization of interleaved video segments. Additionally, \textit{N-Back} probes symbol grounding and memory capacity across temporal gaps. All evaluations share \textbf{a common design principle}: each is grounded in cognitive psychology theory, instantiated as a controlled video task, and equipped with targeted questions and metrics to quantify specific failure modes. Below, we first introduce the design of each evaluation paradigm in detail and then describe the process of evaluation dataset creation.

\subsection{Evaluation Design}
\subsubsection{Divided Attention: Encoding Concurrent Information}
\label{sec:divided_attention_design}

\begin{figure}[H]
    \vspace{-4mm}
    \centering
    \includegraphics[width=0.9\textwidth]{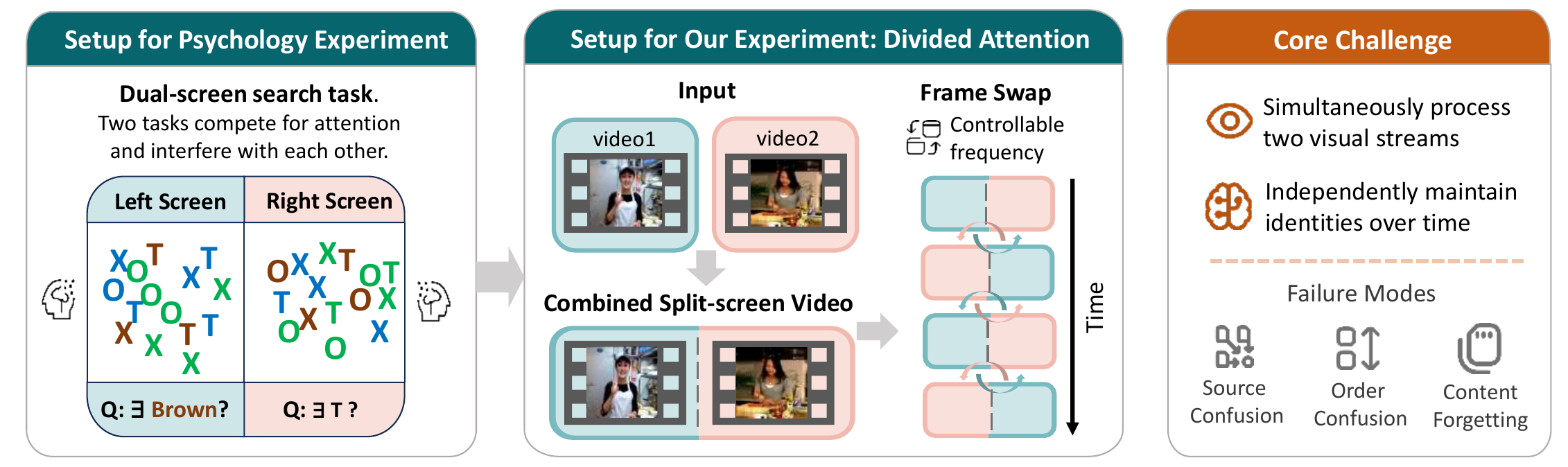}
    \vspace{-3mm}
    \caption{\textbf{Divided Attention}. Split-screen presentation with optional frame swaps.}    
    \label{fig:DA}
    \vspace{-4mm}
\end{figure}

\textbf{Psychological Theory}.
The divided attention paradigm originates from research on \emph{limited attentional resources} and \emph{dual-task processing}~\cite{Kahneman1973-KAHAAE,baddeley1998working}. In classic experiments, participants perform two tasks simultaneously, competing for attentional resources and resulting in reduced encoding quality and impaired memory retention~\cite{Kahneman1973-KAHAAE,craik1996effects,treisman1980feature,treisman1982illusory}.

\textbf{Instantiation in Video Understanding}.
Following this paradigm, we adopt a split-screen configuration where two semantically similar videos are displayed synchronously, as shown in Figure~\ref{fig:DA}. We consider two conditions: (1)~\textit{No swapping}: $V_1$ appears on the left and $V_2$ on the right, evaluating whether the model maintains distinct representations under parallel input. (2)~\textit{Swapping}: the positions of $V_1$ and $V_2$ are swapped 10 times at uniformly spaced timestamps, examining whether the model can track the correspondence between content identity and spatial location.

\textbf{Metrics.}
We construct three types of multiple-choice questions, each targeting a specific failure mode. Each question has one correct option and three distractors of the same error type: (1)~\textit{Source Identification}, where content from the distractor video is erroneously attributed to the target, resulting in source confusion; (2)~\textit{Order Understanding}, where the temporal or logical sequence of events is inaccurately recalled; and (3)~\textit{Content Retention}, where plot points or details from the target video are misremembered or imprecisely recalled.

\subsubsection{Memory Interference: Robustness to Distraction}\label{sec:interference_design}

\begin{figure}[H]
    \vspace{-4mm}   
    \centering
    \includegraphics[width=0.9\textwidth]{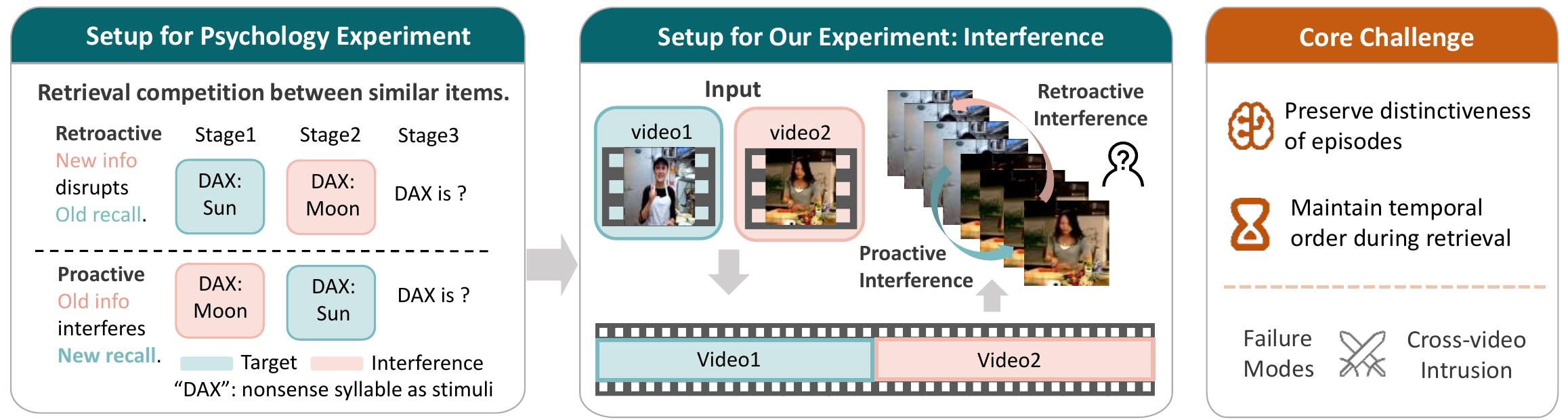}
    \vspace{-3mm}
    \caption{\textbf{Memory Interference}. Proactive interference: earlier learning disrupts later memory. Retroactive interference: later learning impairs earlier memory.}
    \label{fig:INF}
    \vspace{-4mm}
\end{figure}
 
\textbf{Psychological Theory}.
Memory interference theory explains forgetting as competition among similar traces rather than passive decay~\cite{mcgeoch1932forgetting,underwood1957interference}. \textit{Proactive interference} occurs when earlier material disrupts recall of later material, while \textit{retroactive interference} occurs when later material impairs recall of earlier material~\cite{zaromb2006temporal,Robinson1920}. Figure~\ref{fig:INF} (left) illustrates both directions with paired associations.

\textbf{Instantiation in Video Understanding}.
As shown in Figure~\ref{fig:INF}, we concatenate two semantically similar videos and pose questions about one designated target video. To isolate each interference direction, we evaluate both concatenation orders using identical questions targeting the same video. Specifically, in the order [V1,\,V2], asking about V1 tests retroactive interference, as the later video V2 may disrupt recall of the earlier target. In the order [V2,\,V1], asking about the same V1 tests proactive interference, as the earlier video V2 may disrupt encoding of the later target.

\textbf{Metrics.}
We design multiple-choice questions with four options: (1)~the correct answer for the target video, from which we report Accuracy (Acc); (2)~two intrusion options drawn from the competing video, from which we report Intrusion Rate (IR) following~\cite{zaromb2006temporal}, measuring the proportion of responses that select an option from the competing video; and (3)~one unrelated distractor. IR directly quantifies cross-video intrusion.

\subsubsection{Interleaved Events: Temporal Organization}
\label{sec:interleaved_design}

\begin{figure}[H]
    \vspace{-4mm}   
    \centering
    \includegraphics[width=0.9\textwidth]{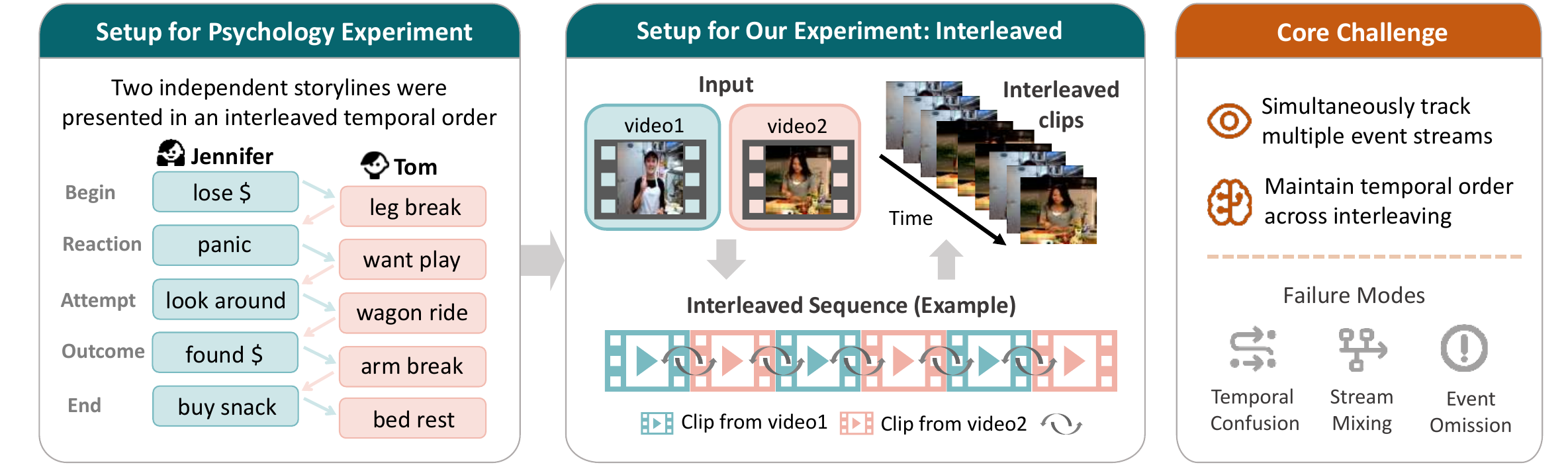}
    \vspace{-3mm}
    \caption{\textbf{Interleaved Events}. Interleaved presentation of video clips from two sources.}
    \label{fig:INTERLEAVED}
    \vspace{-4mm}
\end{figure}

\textbf{Psychological Theory}.
Mandler~\cite{mandler1977remembrance,mandler1978code} demonstrated that, when presented with intermixed storylines, individuals spontaneously recover the underlying event structure rather than following surface presentation order. This paradigm has become a classic test for memory organization.

\textbf{Instantiation in Video Understanding}.
We divide two source videos with each into 10 temporally ordered segments and interleave them into a single stream in alternating order, e.g., $A_1$--$B_1$--$A_2$--$B_2$--$\cdots$--$A_{10}$--$B_{10}$, as shown in Figure~\ref{fig:INTERLEAVED}. To answer correctly, the model must disentangle segments from the same source and recover the internal temporal order of the target video.

\textbf{Metrics.}
We adopt the same three question types as in \S\ref{sec:divided_attention_design}, and add a fourth: ~\textit{False Memory Discrimination}, inspired by the DRM paradigm~\cite{roediger1995creating,deese1959prediction}. 
Here, a fake question that is relevant to video content is presented, and the model should be aware to choose the option indicating that the query does not belong to either video. 

\subsubsection{N-Back: Symbol Grounding and Memory Capacity}
\label{sec:nback_design}
\begin{figure}[H]
    \vspace{-4mm}
    \centering
    \includegraphics[width=0.9\textwidth]{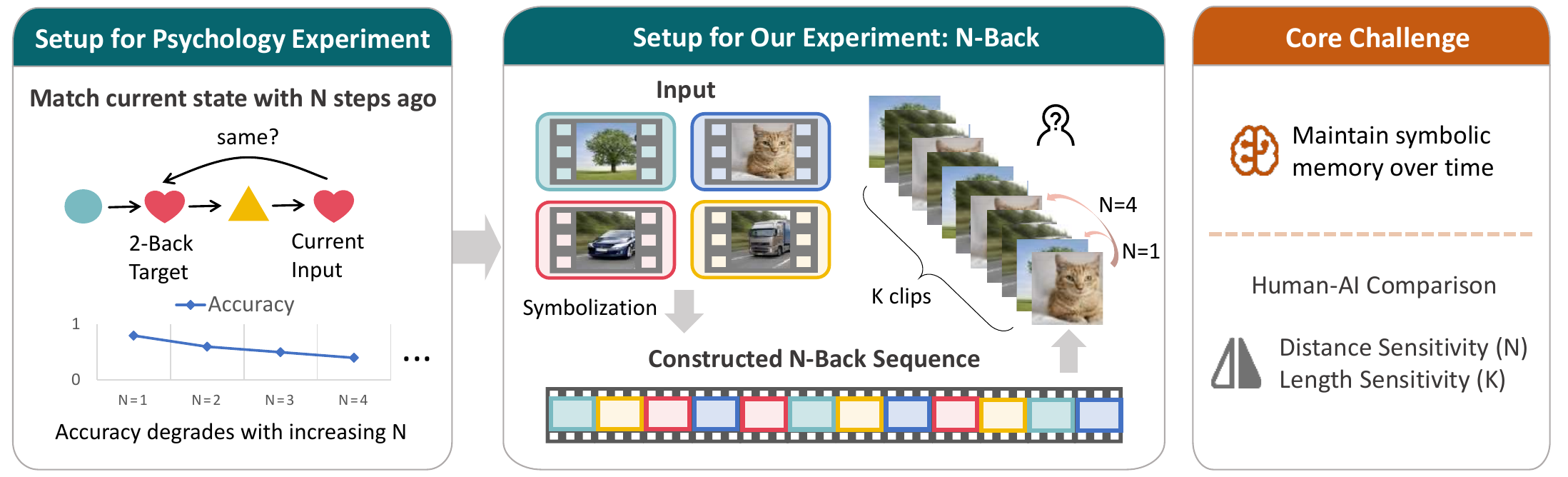}
    \vspace{-3mm}
    \caption{\textbf{N-Back}. Abstracting videos into symbols and comparing them.}
    \label{fig:nback}
    \vspace{-4mm}
\end{figure}

\textbf{Psychological Theory}.
Unlike episodic memory, symbolic memory concerns the ability to abstract events into symbolic representations~\cite{anderson2014human,pylyshyn1973mind}. N-Back tasks present sequences of symbolic stimuli (e.g., letters, digits, or simple shapes) and require participants to decide whether the current stimulus matches the one $N$ steps earlier~\cite{kirchner1958age,owen2005nback}. This match/mismatch structure naturally requires encoding stimuli as abstract symbols before comparison, making the N-Back format well-suited for probing symbolic grounding and memory capacity.

\textbf{Instantiation in Video Understanding}.
We adapt the N-Back paradigm to a multi-video clip sequence setting. As shown in Figure~\ref{fig:nback}, each test sample consists of a sequence of short video clips drawn from different source videos. Two variables control the difficulty: $N$, the lag distance, where the model determines whether the final clip matches the clip $N$ positions earlier on a designated attribute (e.g., scene or action category); and $K$, the sequence length, specifying the total number of video clips presented to the model in a single trial.

\textbf{Metrics.}
The model is asked to decide whether the final clip matches the clip $N$ positions earlier, producing a Yes/No answer.~\textit{Scene} measures whether two clips belong to the same scene or environment category, while~\textit{Action} assesses whether they depict the same type of activity. We report accuracy (Acc) over both attributes across all test samples.

\subsection{Evaluation Dataset Creation}

Our video materials are drawn from five publicly available datasets: HourVideo~\cite{chandrasegaran2024hourvideo}, Video-MME (long-video subset)~\cite{fu2025video}, LVBench~\cite{wang2025lvbench}, InfiniBench (TVQA subset)~\cite{ataallah2025infinibench}, and CrossVid~\cite{li2026crossvid}.
Video pairs are selected based on semantic similarity, as similar content induces stronger memory interference~\cite{underwood1957interference}. 
Questions are automatically generated using Qwen3.5-27B~\cite{qwen35} and refined through manual review. In total, our benchmark comprises 2,403 questions over 451 videos spanning approximately 403 hours. 
Further details on video construction, question generation, and illustrative examples are provided in Appendices~\ref{suppl_statistics}, \ref{suppl_video}, \ref{suppl_qa}, and \ref{suppl_vis}.

%% file: sections/04_experiments.tex
\vspace{-2mm}
\section{Experiments and Results}
\vspace{-2mm}
We evaluate \textbf{two proprietary models} (Gemini-3.1-Pro-Preview~\cite{gemini3pro} and GPT-5.4~\cite{gpt54}), \textbf{five open-weight models} (Qwen3-VL-8B-Instruct~\cite{bai2025qwen3}, Qwen3.5-\{4B, 9B, 27B\}~\cite{qwen35}, and InternVL3.5-8B~\cite{internvl35}), and \textbf{two agentic methods}: VideoLucy~\cite{videolucy}, which adopts Qwen3.5-4B as the VLM and DeepSeek-V4-Pro~\cite{deepseekai2026deepseekv4} as the LLM, and M3-Agent~\cite{m3agent} with its default configuration. We additionally report human performance as a reference. Further details are provided in Appendix~\ref{suppl_experimental_details}.

\input{sections/experiments/4_2_divided_attention}
\input{sections/experiments/4_1_memory_interference}
\input{sections/experiments/4_3_interleaved_events}
\input{sections/experiments/4_5_nback}

%% file: sections/experiments/4_2_divided_attention.tex
\vspace{-2mm}
\subsection{Divided Attention: Encoding Concurrent Information}
\label{sec:divided_attention}

\textbf{Task Recap.} Two similar videos are displayed side by side, with or without periodic left/right swaps, measured by source identification, order understanding, and content retention (\S\ref{sec:divided_attention_design}).

\input{table/divided_attention_swap}

\textbf{Main results}.
Table~\ref{tab:da-swap} shows that divided attention is challenging for existing models. They all exhibit a substantial gap from human performance, with most near chance across metrics except Gemini-3.1-Pro-Preview, indicating that effective dual-stream understanding remains beyond current memory mechanisms.
With frequent swapping, the most prominent drop occurs on \emph{source identification}, while other categories are largely unaffected. This suggests that swapping mainly disrupts source identification rather than order understanding or content retention.

\textbf{Further experiment}. To better understand this failure mode, we examine attention visualizations from representative examples. As shown in Figure~\ref{fig:appendix-selected-attention-merged-preview}, in the single screen format, model attention concentrates on the queried region. However, in a split-screen format, the attention maps become notably more diffused and disorganized. Based on this observation, we hypothesize that the poor performance may stem from attention confusion across concurrent visual streams, preventing the model from selectively attending to the relevant stream.

\IfFileExists{Figs/appendix_attention_selected_merged_preview.pdf}{
\begin{figure*}[t]
\centering
\includegraphics[width=0.98\textwidth]{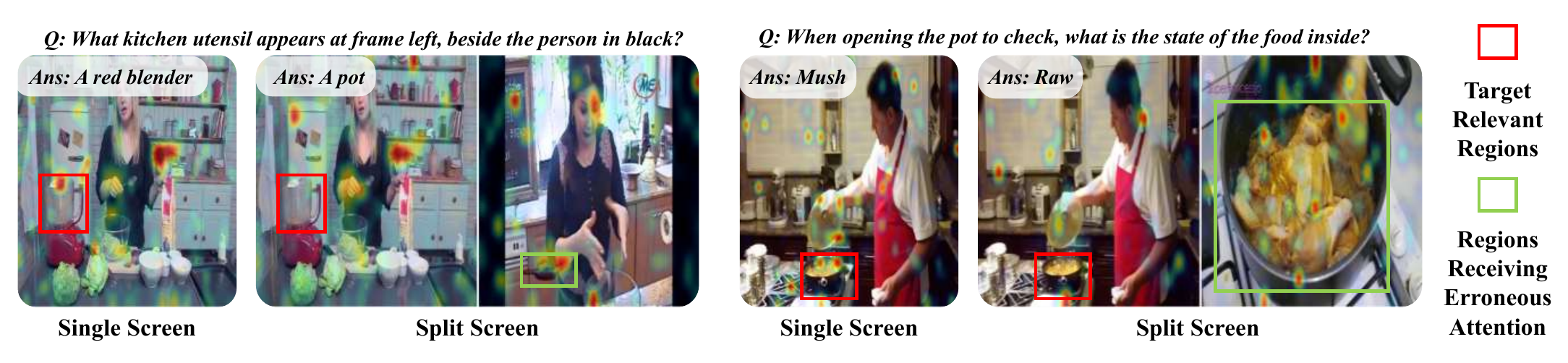}
\vspace{-5mm}
\caption{
\textbf{
Attention shifts induced by split-screen interference}. For each case, the left panel shows the single-video condition, whereas the right panel shows the split-screen condition. 
In the split-screen setting, the question asks specifically about the left video. However, the model's attention is disrupted by the concurrent right video, resulting in erroneous responses.
}
\label{fig:appendix-selected-attention-merged-preview}
\vspace{-6mm}
\end{figure*}
}

\infobox{\textbf{Finding 1}: Existing multi-modal models lack robust memory for parallel tasks, probably due to attention confusion across concurrent visual streams.}

\textbf{Discussion}.
In real-world settings, events often unfold simultaneously, requiring systems to process and reason over multi-view or multi-stream inputs, as in autonomous driving~\cite{jiang2023vad,tian2024drivevlm,liu2025occvla} and household robotics~\cite{intelligence2025pi_,song2026reconvla}. Although existing models perform well on single-video, our experiments suggest they still struggle with parallel streams, multiple objects, and concurrent scenes.

%% file: table/divided_attention_swap.tex
\begin{table}[ht]
\vspace{-2mm}
\centering
\caption{\textbf{Divided Attention}. Accuracy (\%) on three divided attention metrics under the split-screen setting without swaps and with frequent left/right swaps.}
\vspace{-2.5mm}
\label{tab:da-swap}
\small
\begin{adjustbox}{max width=\textwidth}
\begin{tabular}{@{}l ccc ccc@{}}
\toprule
& \multicolumn{3}{c}{\textbf{No swapping}} & \multicolumn{3}{c}{\textbf{Swapping}} \\
\cmidrule(lr){2-4} \cmidrule(lr){5-7}
\textbf{Acc(\%)} & \makecell{Source\\Identification} & \makecell{Order\\Understanding} & \makecell{Content\\Retention} & \makecell{Source\\Identification} & \makecell{Order\\Understanding} & \makecell{Content\\Retention} \\
\midrule
Human & 89.58 & 90.00 & 92.16 & 81.25 {\color{oai-magenta}(\textbf{-}8.33)} & 85.00 {\color{oai-magenta}(\textbf{-}5.00)} & 86.27 {\color{oai-magenta}(\textbf{-}5.89)} \\
\midrule
Random & 25.00 & 25.00 & 25.00 & 25.00 {\color{gray}(0.00)} & 25.00 {\color{gray}(0.00)} & 25.00 {\color{gray}(0.00)} \\
\midrule
\rowcolor{oai-gray-200}
\multicolumn{7}{@{}l}{\textbf{Closed-Source Models}} \\
Gemini-3.1-Pro-Preview & 62.50 & 52.50 & 49.02 & 37.50 {\color{oai-magenta}(\textbf{-}25.00)} & 52.50 {\color{gray}(0.00)} & 56.86 {\color{oai-green}(\textbf{+}7.84)} \\
GPT-5.4 & 27.08 & 35.00 & 47.06 & 35.42 {\color{oai-green}(\textbf{+}8.34)} & 30.00 {\color{oai-magenta}(\textbf{-}5.00)} & 49.02 {\color{oai-green}(\textbf{+}1.96)} \\
\midrule
\rowcolor{oai-gray-200}
\multicolumn{7}{@{}l}{\textbf{Open-Source Agents}} \\
VideoLucy & 16.67 & 42.50 & 37.25 & 14.58 {\color{oai-magenta}(\textbf{-}2.09)} & 25.00 {\color{oai-magenta}(\textbf{-}17.50)} & 39.22 {\color{oai-green}(\textbf{+}1.97)} \\
M3-Agent & 27.08 & 30.00 & 23.53 & 31.25 {\color{oai-green}(\textbf{+}4.17)} & 35.00 {\color{oai-green}(\textbf{+}5.00)} & 23.53 {\color{gray}(0.00)} \\
\midrule
\rowcolor{oai-gray-200}
\multicolumn{7}{@{}l}{\textbf{Open-Source Models}} \\
Qwen3.5-4B & 18.75 & 25.00 & 31.37 & 14.58 {\color{oai-magenta}(\textbf{-}4.17)} & 22.50 {\color{oai-magenta}(\textbf{-}2.50)} & 33.33 {\color{oai-green}(\textbf{+}1.96)} \\
Qwen3-VL-8B-Instruct & 16.67 & 25.00 & 37.25 & 12.50 {\color{oai-magenta}(\textbf{-}4.17)} & 30.00 {\color{oai-green}(\textbf{+}5.00)} & 35.29 {\color{oai-magenta}(\textbf{-}1.96)} \\
InternVL3.5-8B & 29.17 & 37.50 & 33.33 & 25.00 {\color{oai-magenta}(\textbf{-}4.17)} & 40.00 {\color{oai-green}(\textbf{+}2.50)} & 27.45 {\color{oai-magenta}(\textbf{-}5.88)} \\
Qwen3.5-9B & 35.42 & 25.00 & 25.49 & 18.75 {\color{oai-magenta}(\textbf{-}16.67)} & 30.00 {\color{oai-green}(\textbf{+}5.00)} & 13.73 {\color{oai-magenta}(\textbf{-}11.76)} \\
Qwen3.5-27B & 41.67 & 25.00 & 35.29 & 27.08 {\color{oai-magenta}(\textbf{-}14.59)} & 32.50 {\color{oai-green}(\textbf{+}7.50)} & 35.29 {\color{gray}(0.00)} \\
\bottomrule
\vspace{-6mm}
\end{tabular}
\end{adjustbox}
\end{table}

%% file: sections/experiments/4_1_memory_interference.tex
\vspace{-2mm}
\subsection{Memory Interference: Robustness to Distraction}
\label{sec:interference}
\vspace{-2mm}
\textbf{Task Recap.} We concatenate two semantically similar videos (V1 and V2) and ask questions about one designated target video. By swapping the concatenation order — [V1, V2] vs. [V2, V1] — while fixing questions on the same target V1, we isolate retroactive and proactive interference, measured by accuracy and intrusion rate (\S\ref{sec:interference_design}).
\input{table/memory_interference}

\textbf{Main results}.
As shown in Table~\ref{tab:interference}, most models achieve low accuracy, indicating that memory interference poses a significant challenge. 
Further, humans demonstrate a clear asymmetry between proactive and retroactive interference ($\Delta = 20.00\%$), yet models exhibit a small delta between two conditions. This suggests that the models differ from humans in memory mechanism where later information tends to overwrite earlier memories for humans. 
Notably, intrusion rates are high across most models, and thus most errors come from the interference of competing video. This indicates that models struggle to resist interference from semantically similar content.

\newpage

\begin{wrapfigure}{l}{0.48\columnwidth}
    \vspace{-0.4\baselineskip}
    \centering
    \includegraphics[width=\linewidth]{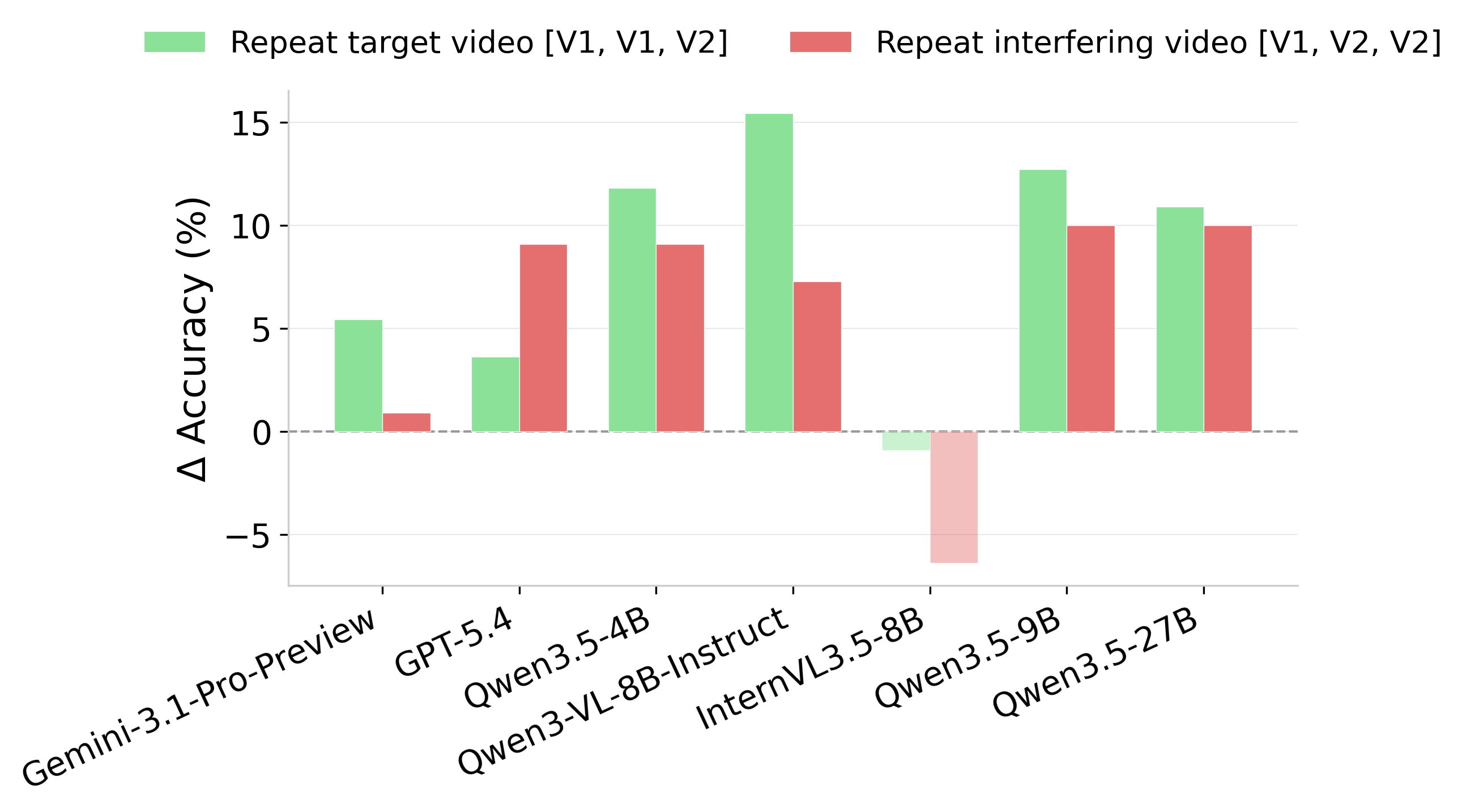}
    \vspace{-8mm}
    \caption{\textbf{Video repetition improves accuracy under interference.} Repeating either the target or interfering video yields performance gains, suggesting repetition as a promising strategy for enhancing model memory.}
    \label{fig:mi-repetition-delta-summary}
    \vspace{-4.5mm}
\end{wrapfigure}

\textbf{Further experiment}.
We test whether repetition strategy can improve robustness to interference. This is done by repeating the target or the interfering video, forming [V1, V1, V2] and [V1, V2, V2] with questions about V1. 
As shown in Figure~\ref{fig:mi-repetition-delta-summary}, both settings surprisingly improve accuracy.
We hypothesize that repetition helps models distinguish the target video from the interfering video. 
Without repetition, causal attention allows later frames to attend to earlier frames, not the reverse; with repetition, the later copy can attend to the earlier occurrence of the same video. 
This gives the model a clearer view of the repeated video, consistent with recent findings~\cite{leviathan2025prompt}.

\infobox{\textbf{Finding 2}: Retroactive interference exceeds proactive interference in humans, whereas both occur comparably in multi-modal models. Further experiments surprisingly find that repeating target or interfering videos can both enhance the understanding of target video.}

\textbf{Discussion}.
Humans exhibit pronounced retroactive interference, whereas most models do not, likely because Transformer attention accesses all visual tokens uniformly regardless of temporal position.
Repetition strategy benefits both humans and models yet through different mechanisms. Humans leverage repetition to reinforce memory anchors~\cite{hintzman1976repetition, cepeda2006distributed}, 
whereas models benefit from the strengthened representations of both target and interfering videos via causal attention.

%% file: table/memory_interference.tex
\begin{table}[ht]
\vspace{-2mm}
\centering
\caption{\textbf{Memory Interference}. Proactive: the first video disrupts recall of the second video. Retroactive: the second disrupts recall of the first. $\Delta$ denotes proactive minus retroactive.}
\vspace{-2.5mm}
\label{tab:interference}
\small
\resizebox{0.72\linewidth}{!}{%
\begin{tabular}{@{}lcccccc@{}}
\toprule
& \multicolumn{3}{c}{\textbf{Accuracy (\%, $\uparrow$)}} & \multicolumn{3}{c}{\textbf{Intrusion Rate (\%, $\downarrow$)}} \\
\cmidrule(lr){2-4} \cmidrule(lr){5-7}
& \textbf{Proactive} & \textbf{Retroactive} & \textbf{$\Delta$} & \textbf{Proactive} & \textbf{Retroactive} & \textbf{$\Delta$} \\
\midrule
Human & 94.55 & 74.55 & 20.00 & 3.64 & 20.00 & -16.36 \\
\midrule
Random & 25.00 & 25.00 & 0.00 & 50.00 & 50.00 & 0.00 \\
\midrule
\rowcolor{oai-gray-200}
\multicolumn{7}{@{}l}{\textbf{Closed-Source Models}} \\
Gemini-3.1-Pro-Preview & 63.64 & 54.55 & 9.09 & 23.64 & 30.91 & -7.27 \\
GPT-5.4 & 43.64 & 40.00 & 3.64 & 43.64 & 34.55 & 9.09 \\
\midrule
\rowcolor{oai-gray-200}
\multicolumn{7}{@{}l}{\textbf{Open-Source Agents}} \\
VideoLucy & 29.09 & 43.64 & -14.55 & 43.64 & 34.55 & 9.09 \\
M3-Agent & 43.64 & 36.36 & 7.28 & 40.00 & 34.55 & 5.45 \\
\midrule
\rowcolor{oai-gray-200}
\multicolumn{7}{@{}l}{\textbf{Open-Source Models}} \\
Qwen3.5-4B & 29.09 & 38.18 & -9.09 & 45.45 & 38.18 & 7.27 \\
Qwen3-VL-8B-Instruct & 25.45 & 29.09 & -3.64 & 54.55 & 52.73 & 1.82 \\
InternVL3.5-8B & 52.73 & 49.09 & 3.64 & 32.73 & 41.82 & -9.09 \\
Qwen3.5-9B & 29.09 & 38.18 & -9.09 & 50.91 & 41.82 & 9.09 \\
Qwen3.5-27B & 45.45 & 40.00 & 5.45 & 40.00 & 43.64 & -3.64 \\
\bottomrule
\vspace{-9mm}
\end{tabular}
}
\end{table}

%% file: sections/experiments/4_3_interleaved_events.tex
\vspace{-2mm}
\subsection{Interleaved Events: Temporal Organization}
\vspace{-2mm}
\label{sec:interleaved}
\textbf{Task Recap.} Segments from two videos are interleaved into a single stream, measured by source identification, order understanding, content retention, and false memory discrimination (\S\ref{sec:interleaved_design}).

\begin{wrapfigure}{l}{0.48\columnwidth}
    \vspace{-4mm}
    \vspace{-0.4\baselineskip}
    \centering
    \includegraphics[width=\linewidth]{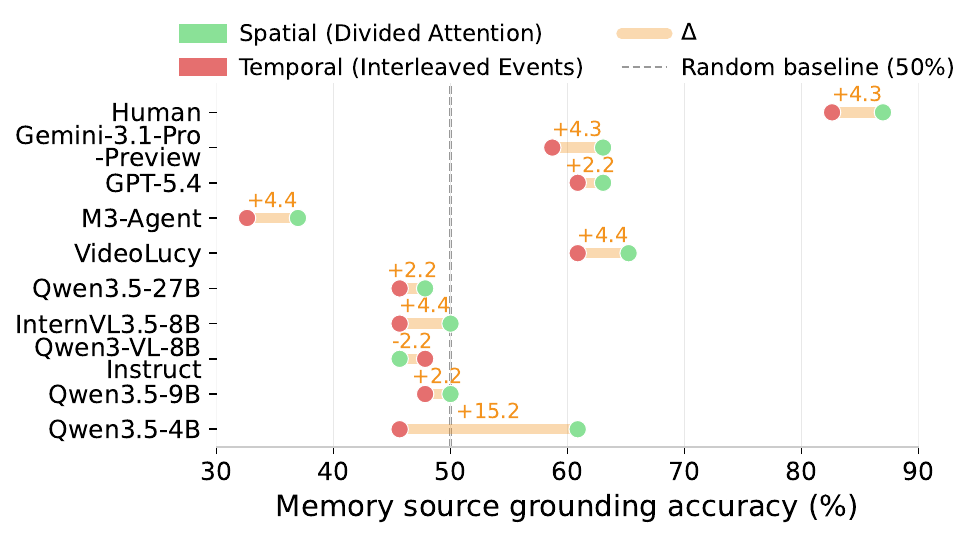}
    \vspace{-8mm}
    \caption{\textbf{Spatial source grounding outperforms temporal source grounding.} Spatial source uses the split-screen format with frequent left/right swaps (\S\ref{sec:divided_attention}); temporal source uses the interleaved format (\S\ref{sec:interleaved}).}
    \label{fig:source-mem}
    \vspace{-6mm}
\end{wrapfigure}

\input{table/interleaved_hint}

\textbf{Main results}.
As shown in Table~\ref{tab:ie-hint}, humans substantially outperform all models across all four question types. These results demonstrate that reorganizing temporally interleaved events remains a significant challenge.
Agentic methods show no clear advantage, suggesting that rule-based memory strategies are insufficient for handling temporal interleaving.
Notably, most models achieve below the 25\% random baseline on false memory discrimination, revealing severe hallucination.

\textbf{Further experiment}.
To further examine the ability of memory source grounding~\cite{johnson1993source, schacter1984retrieval}, we compare grounding accuracy under spatial (split-screen with frequent left/right swaps, \S\ref{sec:divided_attention}) versus temporal (interleaved, \S\ref{sec:interleaved}) conditions.
As shown in Figure~\ref{fig:source-mem}, spatial source grounding generally yields higher accuracy than temporal source grounding, where many models even fall below the random baseline. 
These results suggest that for both humans and models, accurately grounding the temporal source is more difficult than grounding the spatial source.

\infobox{\textbf{Finding 3}: Compared to human memory, multi-modal models are less capable of organizing temporally interleaved information. Further analysis reveals that memory source grounding along the spatial dimension is consistently stronger than along the temporal dimension.}

\textbf{Discussion}.
Models exhibit stronger spatial source grounding than temporal source grounding, mirroring an asymmetry observed in human cognition~\cite{torres2025neurophysiological, pathman2018space} and AI research~\cite{upadhyay2025time}. This suggests that temporal memory organization is inherently more challenging. One potential direction is building models to better capture sequential relationships across events.

%% file: table/interleaved_hint.tex
\vspace{-2mm}
\begin{table}[ht]
\centering
\caption{\textbf{Interleaved Events}. Accuracy (\%) on four interleaved reconstruction metrics.}
\vspace{-2.5mm}
\label{tab:ie-hint}
\small
\begin{adjustbox}{max width=0.81\textwidth}
\begin{tabular}{@{}l cccc@{}}
\toprule
\textbf{Acc(\%)} & Source Identification & Order Understanding & Content Retention & \makecell{False Memory\\Discrimination} \\
\midrule
Human & 75.95 & 80.00 & 83.64 & 82.11 \\
\midrule
Random & 25.00 & 25.00 & 25.00 & 25.00 \\
\midrule
\rowcolor{oai-gray-200}
\multicolumn{5}{@{}l}{\textbf{Closed-Source Models}} \\
Gemini-3.1-Pro-Preview & 43.04 & 50.00 & 49.09 & 26.32 \\
GPT-5.4 & 43.04 & 40.00 & 47.27 & 7.37 \\
\midrule
\rowcolor{oai-gray-200}
\multicolumn{5}{@{}l}{\textbf{Open-Source Agents}} \\
VideoLucy & 30.38 & 23.33 & 43.64 & 40.00 \\
M3-Agent & 27.85 & 40.00 & 21.82 & 15.79 \\
\midrule
\rowcolor{oai-gray-200}
\multicolumn{5}{@{}l}{\textbf{Open-Source Models}} \\
Qwen3.5-4B & 30.38 & 20.00 & 41.82 & 23.16 \\
Qwen3-VL-8B-Instruct & 21.52 & 23.33 & 30.91 & 3.16 \\
InternVL3.5-8B & 25.32 & 26.67 & 41.82 & 1.05 \\
Qwen3.5-9B & 26.58 & 40.00 & 25.45 & 7.37 \\
Qwen3.5-27B & 39.24 & 33.33 & 34.55 & 3.16 \\
\bottomrule
\vspace{-6mm}
\end{tabular}
\end{adjustbox}
\end{table}

%% file: sections/experiments/4_5_nback.tex
\vspace{-2mm}
\subsection{N-Back: Symbol Grounding and Memory Capacity}
\vspace{-2mm}
\label{sec:nback}

\textbf{Task Recap.} A sequence of $K$ short video clips is presented, and the model determines whether the final clip matches the one $N$ positions earlier on a designated attribute, measured by accuracy on scene and action matching (\S\ref{sec:nback_design}).

\vspace{-4mm}
\begin{figure}[H]
    \centering
    \includegraphics[width=0.75\textwidth]{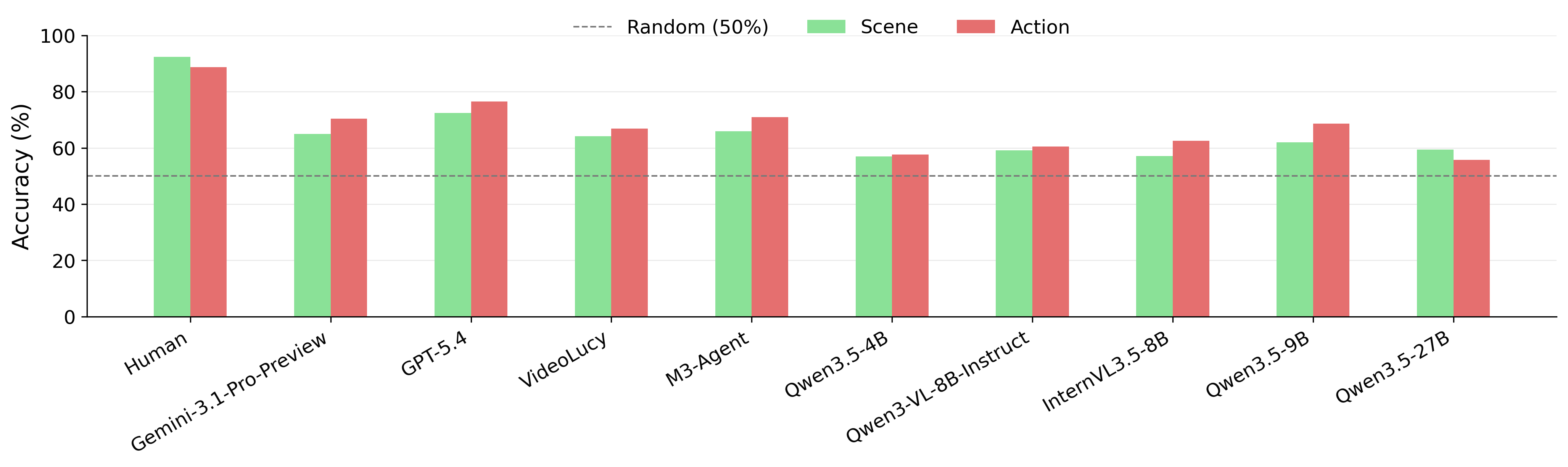}
    \vspace{-4mm}
    \caption{\textbf{Overall accuracy on the N-Back task.} Performance of each model and human under two symbolic attributes (scene and action), averaged over all $K$ and $N$ configurations.}
    \vspace{-5mm}
\end{figure}

\textbf{Main results}. 
Existing multi-modal models substantially lag behind humans, with many only slightly exceeding the random baseline. Among them, GPT-5.4 achieves the best performance. Interestingly, humans recall scene attributes more accurately than action attributes, whereas most models show the opposite pattern, with action accuracy being higher than scene accuracy.

\begin{wrapfigure}{l}{0.48\columnwidth}
    \vspace{-0.4\baselineskip}
    \centering
    \vspace{-4mm}
    \includegraphics[width=0.48\textwidth]{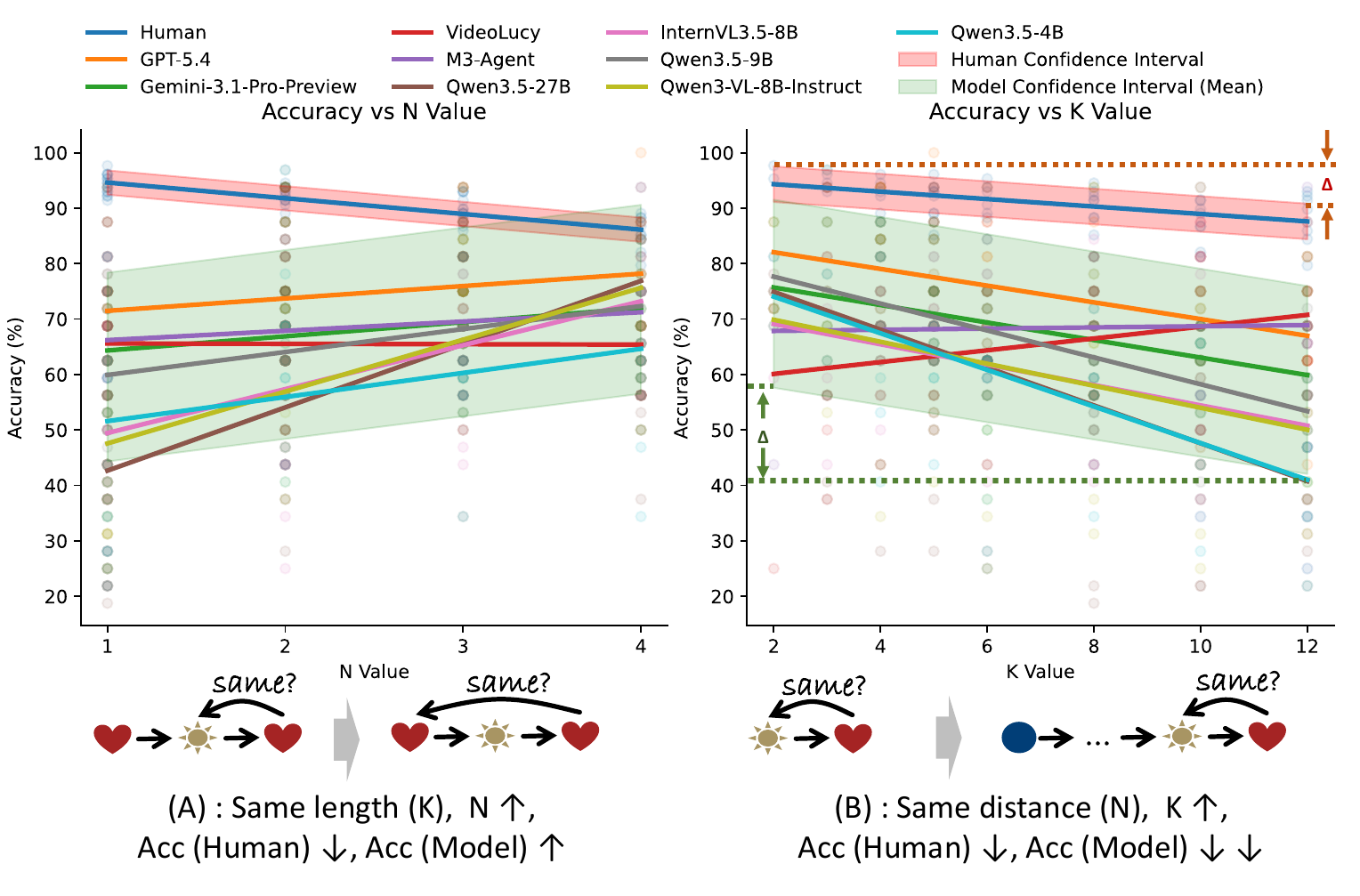}
    \vspace{-8mm}
    \caption{\textbf{Effects of $N$ and $K$ on accuracy.} Points show per-model accuracy under different $(N, K)$ settings, with linear fits for each model. The colored filled regions indicate $\pm 1$ standard deviation around the fit lines.}
    \label{fig:N-back-2}
    \vspace{-6mm}
\end{wrapfigure}

\textbf{Further experiment}.
As shown in Figure~\ref{fig:N-back-2}, behavior discrepancy emerges between humans and models. 
For humans, monotonic decline of accuracy is observed with increasing $N$, reflecting capacity limit. When increasing $K$, accuracy decreases modestly, demonstrating an ability to discard irrelevant information. For example, when $N=2$ and $K=9$, the first six video clips are no longer relevant to the final decision and thus can be discarded.
In contrast, model accuracy remains flat or even improves as $N$ increases, likely due to the Transformer architecture that retrieves temporally distant clips through global attention. However, accuracy drops sharply with increasing $K$, suggesting that models struggle to filter out irrelevant information from memory.

\infobox{\textbf{Finding 4}: Multi-modal models lag far behind humans in symbolic memory. Unlike humans, models do not decay when increasing temporal gap (N), yet degrade largely with increasing number of total symbols (K), revealing a fundamental inability to filter irrelevant memory.}

\textbf{Discussion}. In the N-Back task, humans typically maintain only recent items in working memory while gradually forgetting earlier ones. By contrast, current models retain all past inputs at a similar level of accessibility due to the attention mechanism. We hypothesize that introducing an appropriate forgetting mechanism could help multi-modal models overcome the limitations of symbolic memory, complementing recent explorations in AI research~\cite{hu2023chatdb, wang2024symbolic}.

%% file: sections/06_conclusion.tex
\vspace{-2mm}
\section{Conclusion}
\vspace{-2mm}
In this work, we introduce $M^3Eval$, the first benchmark for systematically measuring multi-modal memory across different dimensions. $M^3Eval$ is grounded in cognitive psychology and instantiated through orchestrated video tasks, moving beyond conventional video understanding benchmarks to probe memory mechanisms critical for multi-modal models.
Our experiments reveal consistent weaknesses and meaningful characteristics across models, pointing to several future directions: \textit{(1)} refining attention mechanisms to better handle parallel streams; \textit{(2)} leveraging repetition strategy to mitigate interference between similar memory traces; \textit{(3)} strengthening temporal source grounding, which substantially lags behind spatial grounding; and \textit{(4)} improving symbolic memory to support abstraction and filtering of task-irrelevant memory. We hope that $M^3Eval$ will serve as a diagnostic tool for future research and motivate the development of multi-modal systems equipped with robust, structured, and human-aligned memory capabilities.

%% file: suppl_content_new.tex
\input{sections/suppl_benchmark_statistics}

%% file: sections/suppl_benchmark_statistics.tex
\section{Benchmark Scale Statistics}
\label{suppl_statistics}

\subsection{Question Count}

The full $M^3Eval$ benchmark comprises 2,403 questions, organized into two parts that target different dimensions of multi-modal memory.

\paragraph{Non-N-Back Questions.}
This part contains 739 questions, evaluating divided attention, memory interference, and interleaved events (\S\ref{sec:divided_attention}, \ref{sec:interference}, \ref{sec:interleaved}). We construct the question-answer pairs from 451 videos sourced from six public long-video understanding datasets. Table~\ref{tab:appendix-benchmark-by-dataset} details the distribution of questions across these datasets.

\begin{table}[H]
\centering
\small
\setlength{\tabcolsep}{12pt}
\caption{Composition of the non-N-Back portion of $M^3Eval$ by source dataset.}
\begin{tabular}{lrr}
\toprule
Dataset & Questions & Videos \\
\midrule
CrossVid-CC        & 138 & 85 \\
CrossVid-NC        & 96  & 70 \\
HourVideo          & 100 & 54 \\
InfiniBench-TVQA   & 184 & 95 \\
LVBench            & 102 & 71 \\
Video-MME-L        & 119 & 76 \\
\midrule
Total (Non-N-Back) & 739 & 451 \\
\bottomrule
\end{tabular}
\vspace{0.8em}
\label{tab:appendix-benchmark-by-dataset}
\end{table}

\paragraph{N-Back Questions.}
This part includes 1,664 questions in an N-Back format. We generate them from 64 carefully selected 12-clip sequence instances, evenly split into 32 for the \textit{action} attribute and 32 for the \textit{scene} attribute. Each instance yields 26 valid $K\times N$ combinations, ensuring comprehensive coverage across different memory loads and temporal gaps.

\subsection{Video Duration}

\begin{figure}[H]
\centering
\includegraphics[width=0.82\textwidth]{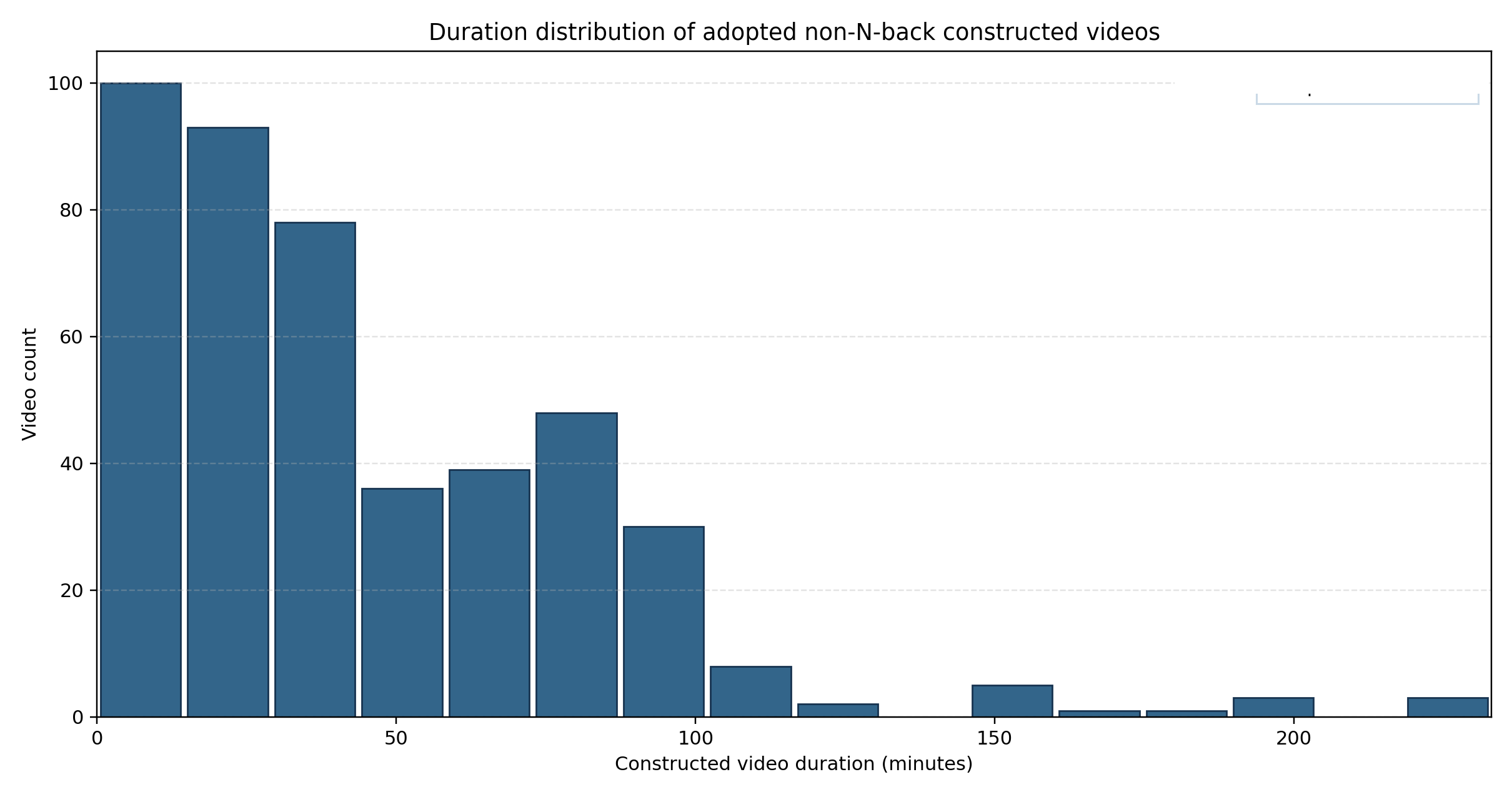}
\caption{Duration histogram for the source videos in the non-N-Back portion of $M^3Eval$. The distribution shows the range of video lengths used in the divided attention, memory interference, and interleaved events tasks.}
\label{fig:appendix-duration-distribution}
\end{figure}

Figure~\ref{fig:appendix-duration-distribution} shows the duration distribution of the 451 source videos used for the non-N-Back tasks. Each clip used in the N-Back tasks is trimmed from these source videos.

%% file: Appendix/qa_construction.tex
\definecolor{PromptBg}{HTML}{F4F1EA} 
\definecolor{PromptFrame}{HTML}{D1C9B8}
\definecolor{PromptText}{HTML}{333333}

\newtcolorbox{promptbox}[1][]{
    enhanced,
    breakable, 
    colback=PromptBg,
    colframe=PromptFrame,
    coltitle=black,
    fonttitle=\scshape\small,
    fontupper=\ttfamily\color{PromptText}\small, 
    boxrule=1pt, 
    arc=4pt, 
    drop shadow=black!5!white, 
    title=#1, 
    left=8pt,
    right=8pt,
    top=8pt,
    bottom=8pt,
    before skip=10pt,
    after skip=10pt
}

\section{Video Construction Details}
\label{suppl_video}

We source video materials from five public datasets and benchmarks: HourVideo~\cite{chandrasegaran2024hourvideo}, Video-MME (long-video subset)~\cite{fu2025video}, LVBench~\cite{wang2025lvbench}, InfiniBench (TVQA subset)~\cite{ataallah2025infinibench}, and CrossVid~\cite{li2026crossvid}. 
These datasets cover egocentric daily activities, diverse web videos, TV episodes, cooking tutorials, and movies, ensuring broad coverage of real-world video scenarios.
We construct video pairs following a semantic similarity-first principle. Within each dataset, videos with similar topics, scenes, or narrative structures are paired. This design is motivated by findings in cognitive psychology that memory interference is strongest between semantically similar materials~\cite{underwood1957interference}.

\section{QA Construction Details}
\label{suppl_qa}
This appendix complements the QA construction pipeline introduced in the main text. We design two separate pipelines for the different memory dimensions targeted by our benchmark.
The first, non-N-Back QA construction, covers Divided Attention, Memory Interference, Interleaved Events, and source-memory judgment sub-tasks.
The second, N-Back QA construction, targets tracking abstract symbolic attributes (scene or action categories) over a video stream.
\subsection{Non-N-Back QA Construction}
This pipeline generates questions through a multi-stage process:
(1) video segmentation;
(2) hierarchical description extraction;
(3) model-based question generation using predefined prompts;
and (4) manual filtering and verification. We manually filter out controversial and potentially composite scenarios, and ensure that the labels within each group are free from interference and ambiguity.
\subsubsection{Video Segmentation}
\label{app:segmentation_description}
Instead of processing videos end-to-end, we first segment them into short, localized units by sampling frames and grouping them into local segments. These segments serve as the basic units for all subsequent description extraction and question generation steps.
\subsubsection{Hierarchical Description Extraction}
For each video, we extract structured evidence at both the local and global levels to prevent the language model from hallucinating or relying on unstructured information. Each segment is described using a predefined six-key schema, as summarized in Table~\ref{tab:caption_keys}.

\begin{table}[h]
\centering
\small
\caption{Structured caption schema for video segments.}
\begin{tabular}{@{}lp{10cm}@{}}
\toprule
\textbf{Key} & \textbf{Description} \\
\midrule
\texttt{main\_storyline} & Plot progression, operational step, or event stage. \\
\texttt{spatial\_relation\_binding} & Layout of people, objects, text, or markers relative to local screen space (left/right, above/below, on-screen region). \\
\texttt{short\_term\_action\_state} & Transient action states or instantaneous changes (open/close, pick up/put down, brief flashes). \\
\texttt{tool\_prop} & Tools, containers, handheld objects, or props relevant to the current operation. \\
\texttt{fine\_visual\_attribute} & Colors, textures, materials, shapes, accessories, or other fine-grained appearance details. \\
\texttt{text\_symbol} & Text, numbers, icons, labels, logos, or other symbolic information. \\
\bottomrule
\end{tabular}
\vspace{1ex}
\label{tab:caption_keys}
\end{table}

\subsubsection{Question Generation Prompts}
Below, we show the prompts fed to the model during the question generation stage for each task and failure mode.

\clearpage

\begin{promptbox}[Video Caption Generation]
You are a professional video evidence describer preparing context for difficult memory questions over two highly similar videos.

You will receive a small batch of nearby frames from one video. Describe only grounded, directly visible evidence from these frames.

Return STRICT JSON only with exactly these 6 keys:
\texttt{main\_storyline},
\texttt{spatial\_relation\_binding},
\texttt{short\_term\_action\_state},
\texttt{tool\_prop},
\texttt{fine\_visual\_attribute},
\texttt{text\_symbol}.

Write every value as one grounded English string.
If a category is not supported by visible evidence in this frame batch, return an empty string for that key.
Never replace these fields with a single \texttt{caption} key.
Do not add wrapper keys such as \texttt{result}, \texttt{response}, or \texttt{data}.

Output format:

\begin{verbatim}
{
  "main_storyline": "",
  "spatial_relation_binding": "",
  "short_term_action_state": "",
  "tool_prop": "",
  "fine_visual_attribute": "",
  "text_symbol": ""
}
\end{verbatim}

Key definitions:

1.~\texttt{main\_storyline}:
State the local storyline progression or process step shown by this batch.
Also include the directly visible people, entities, objects, actions, and scene context that support that storyline reading.
If the frames imply a short local progression, describe that progression clearly.

2.~\texttt{spatial\_relation\_binding}:
Describe detail-level spatial relations or position binding.
Focus on who or what is to the left/right/front/behind, what is on top of or inside something, where a marking appears, or where text/icons appear within the frame.

3.~\texttt{short\_term\_action\_state}:
Describe short-lived action states or momentary changes visible in this batch.
Focus on brief states such as on/off, open/half-open/closed, held/released, appearing/disappearing, flashing/persistent, steaming/not steaming, and other transient states.

4.~\texttt{tool\_prop}:
Describe visible tools, props, utensils, containers, handheld objects, manipulated small items, or operation-relevant objects.

5.~\texttt{fine\_visual\_attribute}:
Describe subtle visual attributes such as colors, patterns, materials, textures, shapes, markings, accessories, or other precise appearance details.

6.~\texttt{text\_symbol}:
Record readable text, numbers, labels, logos, icons, signs, subtitles, packaging text, or symbolic graphics.
If only part is readable, say what is readable and what is uncertain.

Requirements:
Use objective, concrete English.
Do not speculate beyond what is visible or strongly implied by the shown frames.
Keep each field concise but specific.
Preserve short-lived and discriminative evidence instead of collapsing everything into broad summary.
Output exactly one JSON object and nothing else.
\end{promptbox}

\clearpage

\begin{promptbox}[Source Identification]
\textbf{Role}\quad
You are a professional video memory test question designer.

\textbf{Task}\quad
Two semantically similar videos are shown side by side throughout the full clip. You will receive sparse frame-batch caption contexts for Video~1 and Video~2. Design exactly 1 multiple-choice storyline-reconstruction question asking which narrative paragraph most accurately describes what happened in one target video.

\textbf{Focus}\quad
Build the question around the main storyline, process stages, or high-level event progression of the target video rather than tiny fleeting details.

\textbf{2$\times$2 Combinatorial Design}
\begin{itemize}[nosep,leftmargin=*]
  \item This prompt is for \texttt{source\_confusion}.
  \item Option A must be the only correct paragraph before downstream option shuffle.
  \item Select exactly two non-adjacent manipulable positions, \texttt{slot\_1} and \texttt{slot\_2}, inside one shared storyline scaffold.
  \item For each slot, prepare:
    the real target-video main-stage detail as the \texttt{correct} version;
    a corresponding main-stage detail from the other video as the \texttt{incorrect} version.
  \item The four options must realize the full 2$\times$2 grid:
    \texttt{A}: \texttt{slot\_1=correct}, \texttt{slot\_2=correct};
    \texttt{B}: \texttt{slot\_1=incorrect}, \texttt{slot\_2=correct};
    \texttt{C}: \texttt{slot\_1=correct}, \texttt{slot\_2=incorrect};
    \texttt{D}: \texttt{slot\_1=incorrect}, \texttt{slot\_2=incorrect}.
  \item All three wrong options must stay within the single error family \texttt{source\_confusion}.
\end{itemize}

\textbf{Writing Constraints}
\begin{enumerate}[nosep,leftmargin=*]
  \item The question must ask: ``Which of the following options most accurately describes what happened in Video~X?''
  \item Each option must be one coherent narrative paragraph of about 90--160 words.
  \item All four options must keep nearly the same paragraph structure, writing style, and overall narrative scaffold.
  \item Differences should center on meaningful storyline stages, not trivial wording noise.
  \item Do not turn the paragraph into bullet-like stage labels; keep fluent prose.
  \item The four option strings must all be distinct; never output two identical paragraphs.
\end{enumerate}

\textbf{Self-Check Before Finalizing}
\begin{enumerate}[nosep,leftmargin=*]
  \item All four options are individually coherent and share the same high-level storyline scaffold.
  \item \texttt{A/B/C/D} exactly realize the \texttt{correct-correct / incorrect-correct / correct-incorrect / incorrect-incorrect} grid.
  \item \texttt{option\_role\_by\_letter} stays coarse-grained: \texttt{A=correct}, \texttt{B/C/D=source\_confusion}.
  \item \texttt{storyline\_2x2\_design.slot\_1} and \texttt{slot\_2} each use exact verbatim anchor spans copied from the option text so a checker can verify the expected presence/absence; do not paraphrase or inflect those anchor spans across options.
  \item The two slots are separated in the paragraph and do not collapse into the same local phrase.
  \item \texttt{B} differs from \texttt{A} only at \texttt{slot\_1}, \texttt{C} differs from \texttt{A} only at \texttt{slot\_2}, and \texttt{D} differs from \texttt{A} at both slots.
\end{enumerate}

\end{promptbox}

\clearpage
\begin{promptbox}[Order Understanding]
\textbf{Role}\quad
You are a professional video memory test question designer.

\textbf{Task}\quad
Two semantically similar videos are shown side by side throughout the full clip. You will receive sparse frame-batch caption contexts for Video~1 and Video~2. Design exactly 1 multiple-choice storyline-reconstruction question asking which narrative paragraph most accurately describes what happened in one target video.

\textbf{Focus}\quad
Build the question around the main storyline, process stages, or high-level event progression of the target video rather than tiny fleeting details.

\textbf{2$\times$2 Combinatorial Design}
\begin{itemize}[nosep,leftmargin=*]
  \item This prompt is for \texttt{order\_disruption}.
  \item Option A must be the only correct paragraph before downstream option shuffle.
  \item Select exactly two non-overlapping reversible event pairs, \texttt{slot\_1} and \texttt{slot\_2}, inside one shared storyline scaffold.
  \item For each slot, prepare:
    the true temporal order as the \texttt{correct} version;
    a locally plausible reversed order as the \texttt{incorrect} version.
  \item Represent each slot with the same two short event clauses reused word-for-word across all four options. Form the incorrect version only by swapping the order of those exact clauses.
  \item The four anchor clauses across \texttt{slot\_1} and \texttt{slot\_2} must all be distinct. Do not reuse one event clause in both slots.
  \item In every option, each anchor clause must appear exactly once. Do not duplicate or omit an anchor clause when building a distractor.
  \item Keep each slot as one local two-clause micro-sequence in place. Reverse the two clauses inside the same local sentence window; do not move one anchor clause to another sentence or another slot region.
  \item The four options must realize the full 2$\times$2 grid:
    \texttt{A}: \texttt{slot\_1=correct}, \texttt{slot\_2=correct};
    \texttt{B}: \texttt{slot\_1=incorrect}, \texttt{slot\_2=correct};
    \texttt{C}: \texttt{slot\_1=correct}, \texttt{slot\_2=incorrect};
    \texttt{D}: \texttt{slot\_1=incorrect}, \texttt{slot\_2=incorrect}.
  \item All three wrong options must stay within the single error family \texttt{order\_disruption}.
  \item Both slots must genuinely change somewhere in the final options. Do not leave one slot text unchanged across all four options.
  \item Construct \texttt{A} first. Then create \texttt{B}, \texttt{C}, and \texttt{D} by copying \texttt{A} and swapping only the required slot order in place. Do not rewrite the paragraph from scratch for each letter.
\end{itemize}

\textbf{Writing Constraints}
\begin{enumerate}[nosep,leftmargin=*]
  \item The question must ask: ``Which of the following options most accurately describes what happened in Video~X?''
  \item Each option must be one coherent narrative paragraph of about 90--160 words.
  \item All four options must keep nearly the same paragraph structure, writing style, and overall narrative scaffold.
  \item Differences should center on meaningful storyline stages rather than trivial wording noise.
  \item Do not turn the paragraph into bullet-like stage labels; keep fluent prose.
  \item The four option strings must all be distinct; never output two identical paragraphs.
  \item Build \texttt{slot\_1} and \texttt{slot\_2} as two separate local windows, preferably two separate sentences. Do not let the two slots overlap or share a sentence fragment.
\end{enumerate}

\textbf{Self-Check Before Finalizing}
\begin{enumerate}[nosep,leftmargin=*]
  \item All four options are individually coherent and share the same high-level storyline scaffold.
  \item \texttt{A/B/C/D} exactly realize the \texttt{correct-correct / incorrect-correct / correct-incorrect / incorrect-incorrect} grid.
  \item \texttt{option\_role\_by\_letter} stays coarse-grained: \texttt{A=correct}, \texttt{B/C/D=order\_disruption}.
  \item For each slot, \texttt{correct\_anchors} and \texttt{incorrect\_anchors} must be ordered anchor lists copied verbatim from the option text so a checker can verify the local order; the same anchor wording must be reused word-for-word in every option.
  \item The two slots are separated in the paragraph and the four anchor clauses are all distinct; no anchor phrase is reused across slots.
  \item Every option contains each anchor clause exactly once; no anchor clause is duplicated or omitted.
  \item Each slot is reversed in place inside its own local sentence window; no anchor clause is moved to another sentence or another slot region.
  \item \texttt{B} differs from \texttt{A} only at \texttt{slot\_1}, \texttt{C} differs from \texttt{A} only at \texttt{slot\_2}, and \texttt{D} differs from \texttt{A} at both slots.
  \item Before finalizing, literally check that all four anchor clauses from \texttt{A} still appear once each in \texttt{B}, \texttt{C}, and \texttt{D}.
\end{enumerate}

\end{promptbox}

\begin{promptbox}[Content Retention]
\textbf{Role}\quad
You are a professional video memory test question designer.

\textbf{Task}\quad
Two semantically similar videos are shown side by side throughout the full clip. You will receive sparse frame-batch caption contexts for Video~1 and Video~2. Design exactly 1 multiple-choice storyline-reconstruction question asking which narrative paragraph most accurately describes what happened in one target video.

\textbf{Focus}\quad
Build the question around the main storyline, process stages, or high-level event progression of the target video rather than tiny fleeting details.

\textbf{2$\times$2 Combinatorial Design}
\begin{itemize}[nosep,leftmargin=*]
  \item This prompt is for \texttt{content\_forgetting}.
  \item Option A must be the only correct paragraph before downstream option shuffle.
  \item Select exactly two non-adjacent manipulable positions, \texttt{slot\_1} and \texttt{slot\_2}, inside one shared storyline scaffold.
  \item For each slot, prepare:
    the real target-video main-stage detail as the \texttt{correct} version;
    a plausible but nonexistent main-stage detail as the \texttt{incorrect} version.
  \item The four options must realize the full 2$\times$2 grid:
    \texttt{A}: \texttt{slot\_1=correct}, \texttt{slot\_2=correct};
    \texttt{B}: \texttt{slot\_1=incorrect}, \texttt{slot\_2=correct};
    \texttt{C}: \texttt{slot\_1=correct}, \texttt{slot\_2=incorrect};
    \texttt{D}: \texttt{slot\_1=incorrect}, \texttt{slot\_2=incorrect}.
  \item All three wrong options must stay within the single error family \texttt{content\_forgetting}.
\end{itemize}

\textbf{Writing Constraints}
\begin{enumerate}[nosep,leftmargin=*]
  \item The question must ask: ``Which of the following options most accurately describes what happened in Video~X?''
  \item Each option must be one coherent narrative paragraph of about 90--160 words.
  \item All four options must keep nearly the same paragraph structure, writing style, and overall narrative scaffold.
  \item Differences should center on meaningful storyline stages rather than trivial wording noise.
  \item Do not turn the paragraph into bullet-like stage labels; keep fluent prose.
  \item The four option strings must all be distinct; never output two identical paragraphs.
\end{enumerate}

\textbf{Self-Check Before Finalizing}
\begin{enumerate}[nosep,leftmargin=*]
  \item All four options are individually coherent and share the same high-level storyline scaffold.
  \item \texttt{A/B/C/D} exactly realize the \texttt{correct-correct / incorrect-correct / correct-incorrect / incorrect-incorrect} grid.
  \item \texttt{option\_role\_by\_letter} stays coarse-grained: \texttt{A=correct}, \texttt{B/C/D=content\_forgetting}.
  \item \texttt{storyline\_2x2\_design.slot\_1} and \texttt{slot\_2} each use exact verbatim anchor spans copied from the option text so a checker can verify the expected presence/absence; do not paraphrase or inflect those anchor spans across options.
  \item The two fabricated details never appear in either source video and do not collapse into the same local phrase.
  \item \texttt{B} differs from \texttt{A} only at \texttt{slot\_1}, \texttt{C} differs from \texttt{A} only at \texttt{slot\_2}, and \texttt{D} differs from \texttt{A} at both slots.
\end{enumerate}

\end{promptbox}

\clearpage
\begin{promptbox}[False Memory Discrimination]
You are generating exactly ONE 4-option false-source question for two similar videos named \texttt{video1} and \texttt{video2}.

Canonical type:
\texttt{question\_type\_id}: \texttt{false\_source\_attribution};
\texttt{question\_type}: \texttt{False Source Attribution}.

\textbf{Goal}\quad
Ask about a semantically related event or detail that appears in neither video.
The false event should feel relevant to the two videos, not random.

\textbf{Required Design}
\begin{itemize}[nosep,leftmargin=*]
  \item Use only the provided segment summaries, global summaries, and pair summary as grounding for what is related and what is absent.
  \item The stem detail must be absent from both videos.
  \item Use exactly these 4 option meanings:
    \texttt{Video 1},
    \texttt{Video 2},
    \texttt{Both videos},
    \texttt{Neither video}.
  \item The options themselves must be plain source labels. Do NOT paraphrase them as \texttt{correct video} / \texttt{wrong video}.
  \item Use the source labels directly as answer texts, for example \texttt{A.~Video~1}, \texttt{B.~Video~2}, \texttt{C.~Both videos}, \texttt{D.~Neither video}.
  \item The correct answer must therefore be the \texttt{Neither video} option.
  \item Set \texttt{target\_video} to \texttt{neither}.
\end{itemize}

\end{promptbox}

\begin{promptbox}[Memory Interference]
\textbf{Role}\quad
You are a professional video memory test question designer.

\textbf{Task}\quad
Two semantically similar videos are concatenated into a single continuous video for the model to watch. You will receive sparse frame-caption contexts for each video. Design exactly 1 multiple-choice question that tests whether recall of one video's fine visual attribute is invaded by a similar attribute from the other video.

\textbf{Focus}\quad
Prioritize subtle appearance evidence such as:
colors, patterns, materials, shapes, markings, textures, accessories, or small appearance cues;
details visible only briefly and easy to confuse across similar scenes;
grounded visual attributes rather than broad plot summaries.

\textbf{Runtime Compatibility}
\begin{itemize}[nosep,leftmargin=*]
  \item Treat the provided Video~1 context and Video~2 context as the only source evidence.
  \item Generate exactly ONE 4-option multiple-choice question.
  \item Canonical type:
    \texttt{question\_type\_id}: \texttt{memory\_interference};
    \texttt{question\_type}: \texttt{Memory Interference}.
  \item Set \texttt{target\_video} to \texttt{video1} or \texttt{video2}.
  \item Set \texttt{target\_memory\_ability} to \texttt{memory\_interference\_fine\_visual\_attribute}.
  \item Include \texttt{option\_role\_by\_letter} with exactly one \texttt{correct}, two \texttt{intrusion}, and one \texttt{irrelevant\_distractor}.
  \item Set \texttt{intrusion\_option\_letter} to one of the intrusion letters.
  \item Return STRICT JSON only. No markdown. No explanation.
\end{itemize}

\end{promptbox}

\clearpage
\begin{promptbox}[Source Memory]
\textbf{Role}\quad
You are a professional video memory test question designer.

\textbf{Task}\quad
Two semantically similar videos are presented to the model. You will receive sparse frame-caption contexts for Video~1 and Video~2. Design exactly 1 source-memory multiple-choice question about a short-lived fine visual attribute that truly appeared in only one video.

\textbf{Focus}\quad
Prioritize subtle appearance evidence such as:
colors, patterns, materials, shapes, markings, textures, accessories, or small appearance cues;
details that are visible briefly and are easy to confuse across similar scenes;
grounded visual attributes rather than broad plot summaries.

\textbf{Runtime Compatibility}
\begin{itemize}[nosep,leftmargin=*]
  \item Treat the provided Video~1 context and Video~2 context as the only source evidence.
  \item Generate exactly ONE 2-option multiple-choice question.
  \item Canonical type:
    \texttt{question\_type\_id}: \texttt{source\_memory};
    \texttt{question\_type}: \texttt{Source Memory}.
  \item Use exactly these 2 answer texts:
    \texttt{A.~Video~1},
    \texttt{B.~Video~2}.
  \item Set \texttt{target\_video} to \texttt{video1} or \texttt{video2}.
  \item Set \texttt{target\_memory\_ability} to \texttt{source\_memory\_fine\_visual\_attribute}.
  \item Include \texttt{option\_role\_by\_letter} as a JSON object with exactly:
    \texttt{A}: \texttt{video1\_only};
    \texttt{B}: \texttt{video2\_only}.
  \item Return STRICT JSON only. No markdown. No explanation.
\end{itemize}

\end{promptbox}

\subsubsection{Manual Review and Verification}
All generated candidate questions undergo a rigorous manual review.
We verify the logical consistency of the options, ensure that distractors are plausible but demonstrably incorrect, and revise wording when necessary to eliminate ambiguity.
Only questions that pass this quality check are included in the final benchmark.

\subsection{N-Back QA Construction}
We first uniformly segment each video into clips.
Then, we employ Qwen3.5-27B to annotate each clip with its scene and action attributes via sequential prompting.
Specifically, we process clips in temporal order and maintain a running list of all previously assigned attribute phrases.
At each step, the existing action and scene labels are injected into the prompt, encouraging the model to reuse consistent phrasing for recurring attributes.
A new phrase is introduced only when a genuinely novel action or scene appears.
The annotation prompt is shown below.

\begin{promptbox}[N-Back Attribute Annotation]
You are analyzing a sequence of video segments from the same context. 

Your task is to provide two short phrases for the current segment: 

1. The main Action/Activity.

2. The Scene/Environment.

Current Global Memory for this group:

Existing Actions: [list of action phrases assigned to previous clips]

Existing Scenes: [list of scene phrases assigned to previous clips]

Instructions:

- If the current action/scene matches one in the memory, use the EXACT phrase from the list.

- If it is new, create a concise new phrase (2-4 words).

- OUTPUT FORMAT: Action: [phrase] | Scene: [phrase]
\end{promptbox}

Based on the similarity of these clip-level attributes, we select four groups of videos.
From each group, three clips are sampled and randomly combined to construct N-Back testing video sets.
All generated N-Back probes undergo the same manual review process described above, ensuring that the annotated attributes are accurate and that each probe has a single unambiguous correct answer.

\section{Experimental Details}
\label{suppl_experimental_details}
For frame sampling, we use 0.5~FPS for Gemini-3.1-Pro-Preview and 96 uniform frames for all other models by default, with two exceptions: 144 frames for the repeated-trial experiments in Fig.~\ref{fig:mi-repetition-delta-summary} and 8 frames per clip for the N-Back experiments in Sec.~\ref{sec:nback}.
\textit{For all other settings, we use the defaults}.
All experiments with locally deployed models were conducted on a server equipped with 4 NVIDIA A800 GPUs.
Proprietary models were evaluated through their official APIs.

\section{Example Visualization}
\label{suppl_vis}
Here we present examples for each type of task.
Each example includes: 1) the format in which the video is presented, 2) the task-specific prompt given to the model, and 3) the specific question asked.

\begin{figure}[H]
    \centering
    \includegraphics[width=1\textwidth]{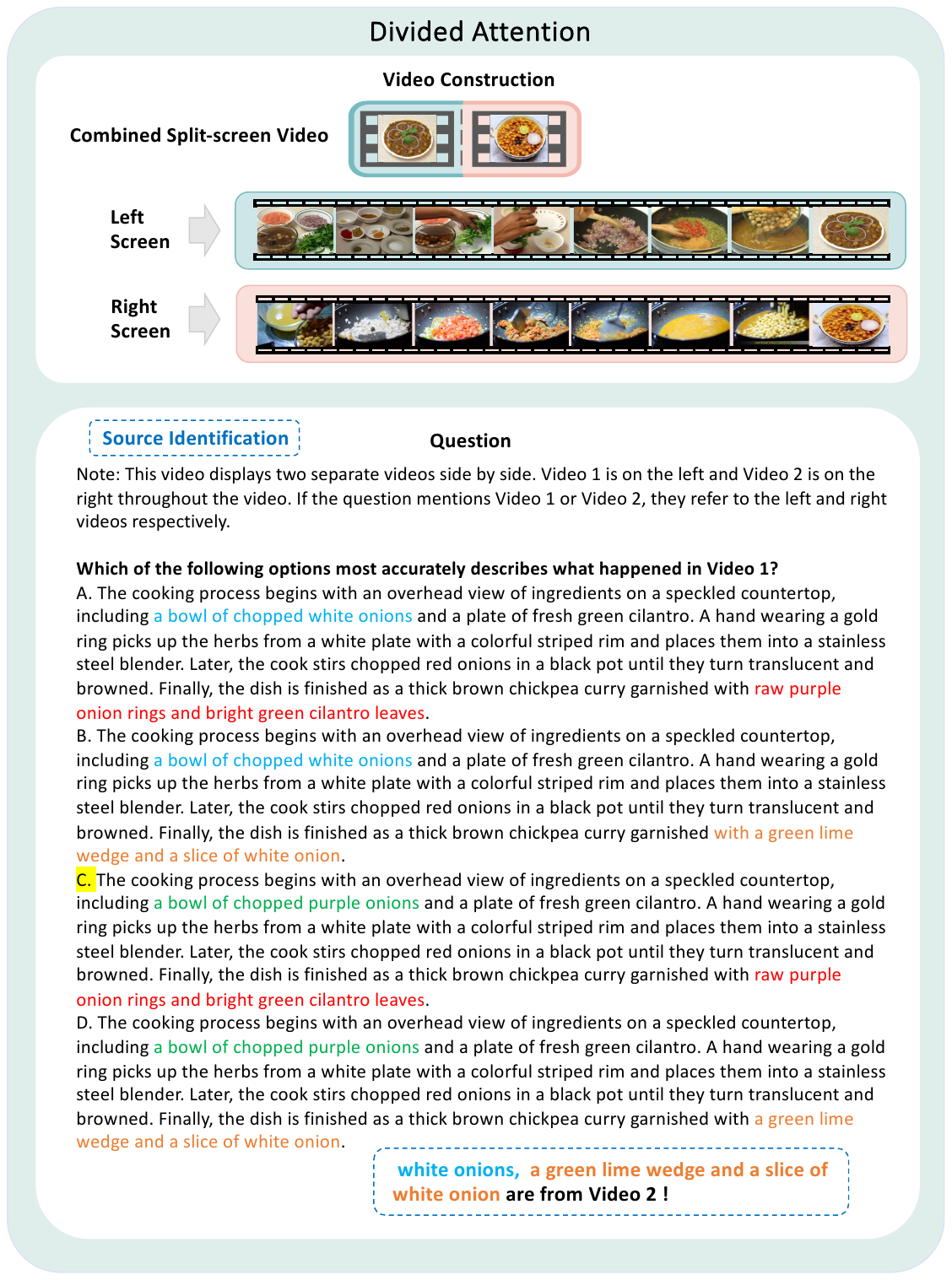}
    \vspace{-20mm}
    \caption{Example of Divided Attention targeting \textit{Source Identification}. The three distractors replace certain content in the target video's narrative with content from the distractor video, while the correct option (highlighted in yellow) faithfully describes only the target video.}
    \label{fig:vis1}
    \vspace{-6mm}
\end{figure}
 
\begin{figure}[H]
    \centering
    \includegraphics[width=1\textwidth]{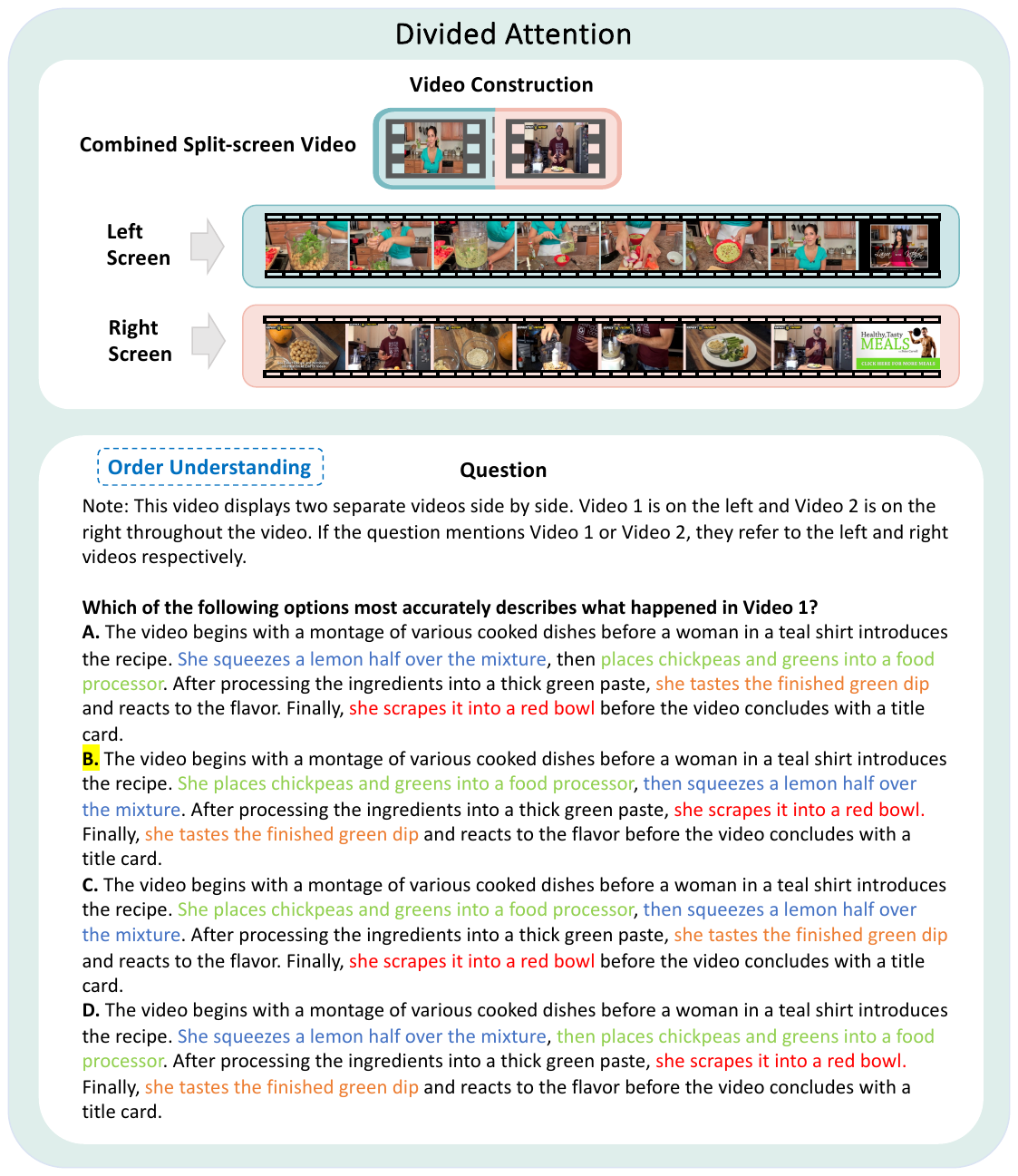}
    \vspace{-32mm}
    \caption{Example of Divided Attention targeting \textit{Order Understanding}. The three distractors swap the temporal or logical sequence of events in the target video's narrative, while the correct option (highlighted in yellow) preserves the original order.}
    \label{fig:vis2}
    \vspace{-6mm}
\end{figure}
 
\begin{figure}[H]
    \centering
    \includegraphics[width=1\textwidth]{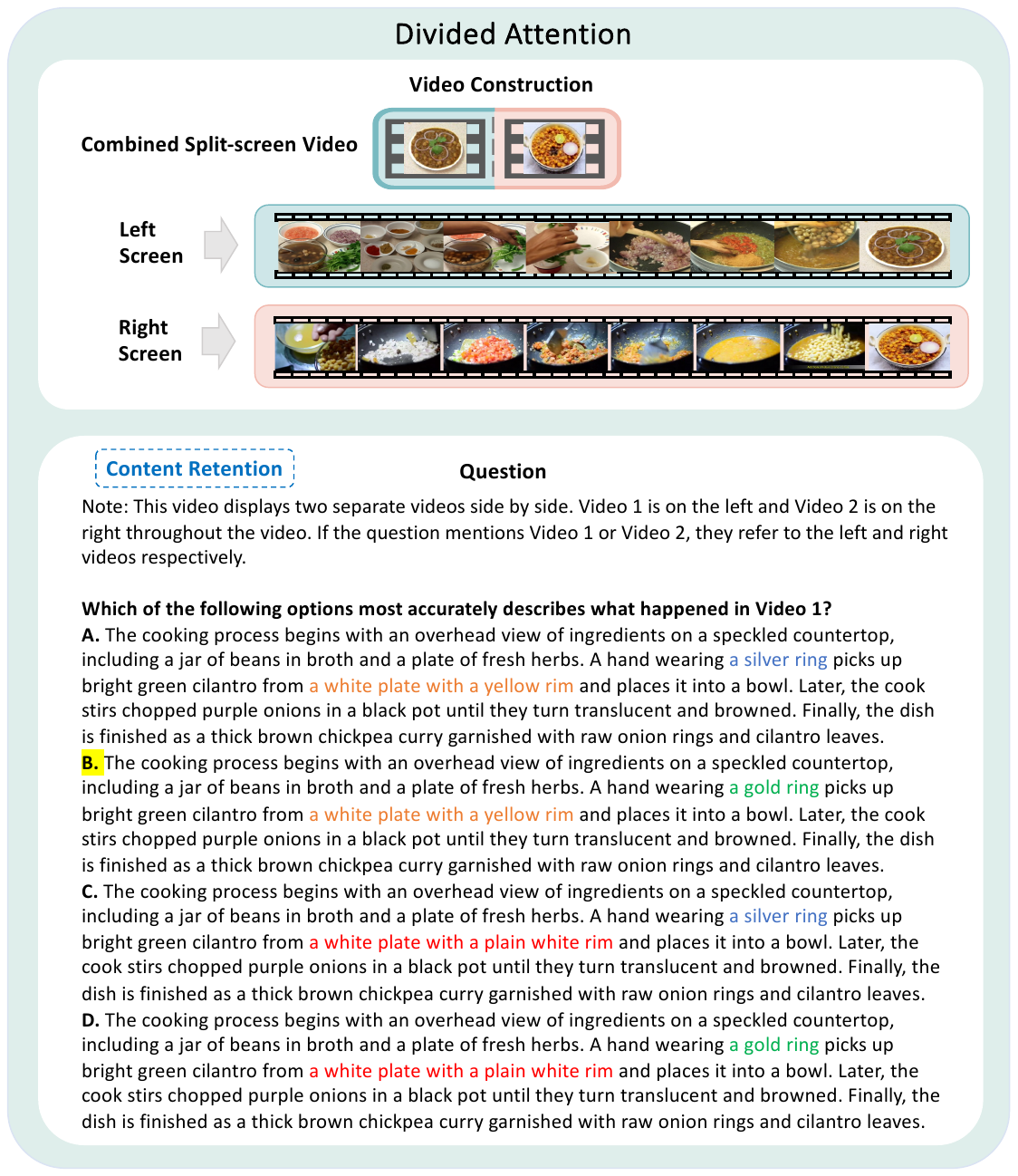}
    \vspace{-32mm}
    \caption{Example of Divided Attention targeting \textit{Content Retention}. The three distractors replace certain content in the target video's narrative with plausible but fabricated content, while the correct option (highlighted in yellow) faithfully describes only the target video.}
    \label{fig:vis3}
    \vspace{-6mm}
\end{figure}
 
\begin{figure}[t]
    \centering
    \includegraphics[width=\textwidth]{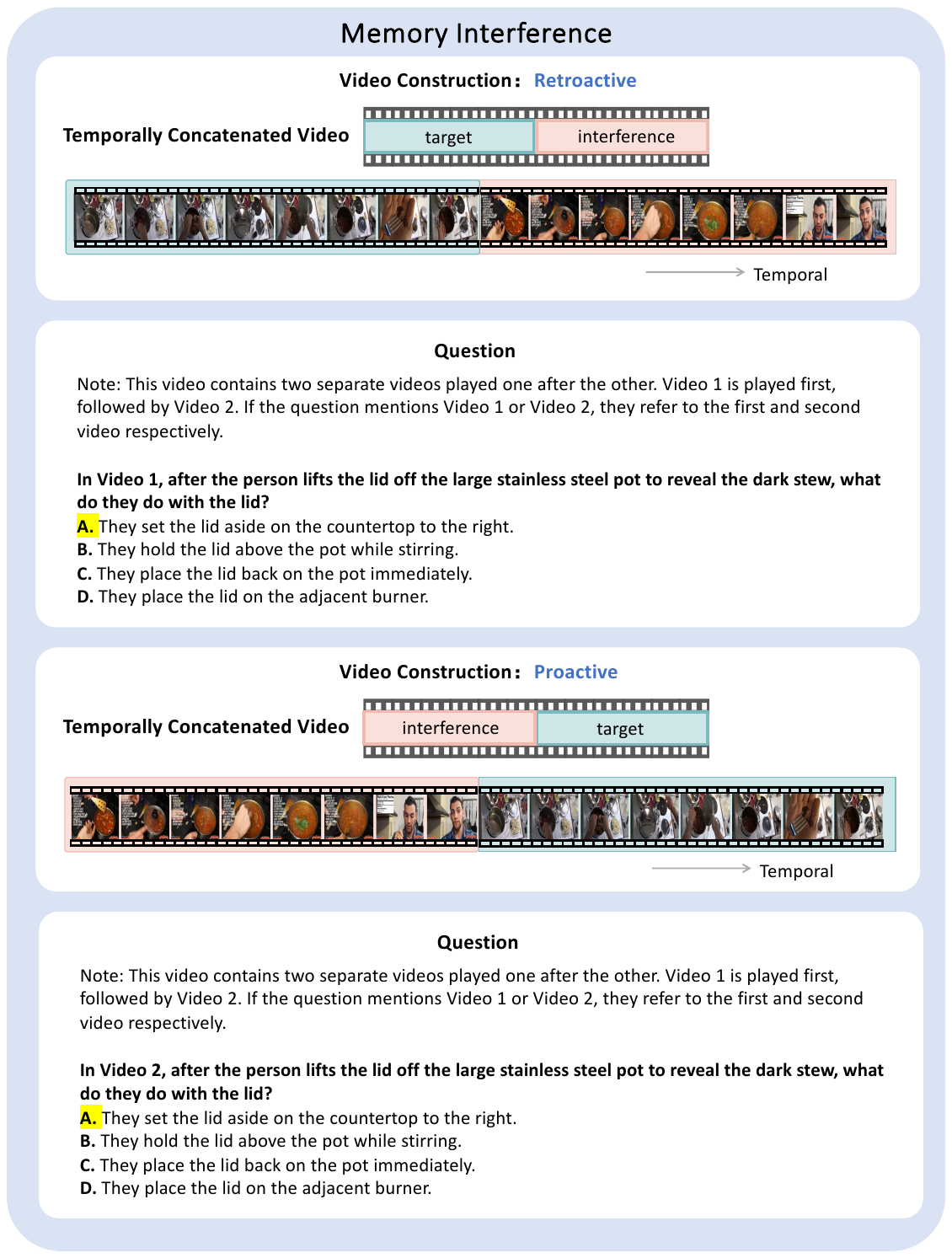}
    \vspace{-22mm}
    \caption{Example of Memory Interference. Each question comprises the correct answer (highlighted in yellow) for the target video, two intrusion options drawn from the interfering video, and one unrelated distractor.}
    \label{fig:vis4}
    \vspace{-6mm}
\end{figure}
 
\begin{figure}[t]
    \centering
    \includegraphics[width=1\textwidth]{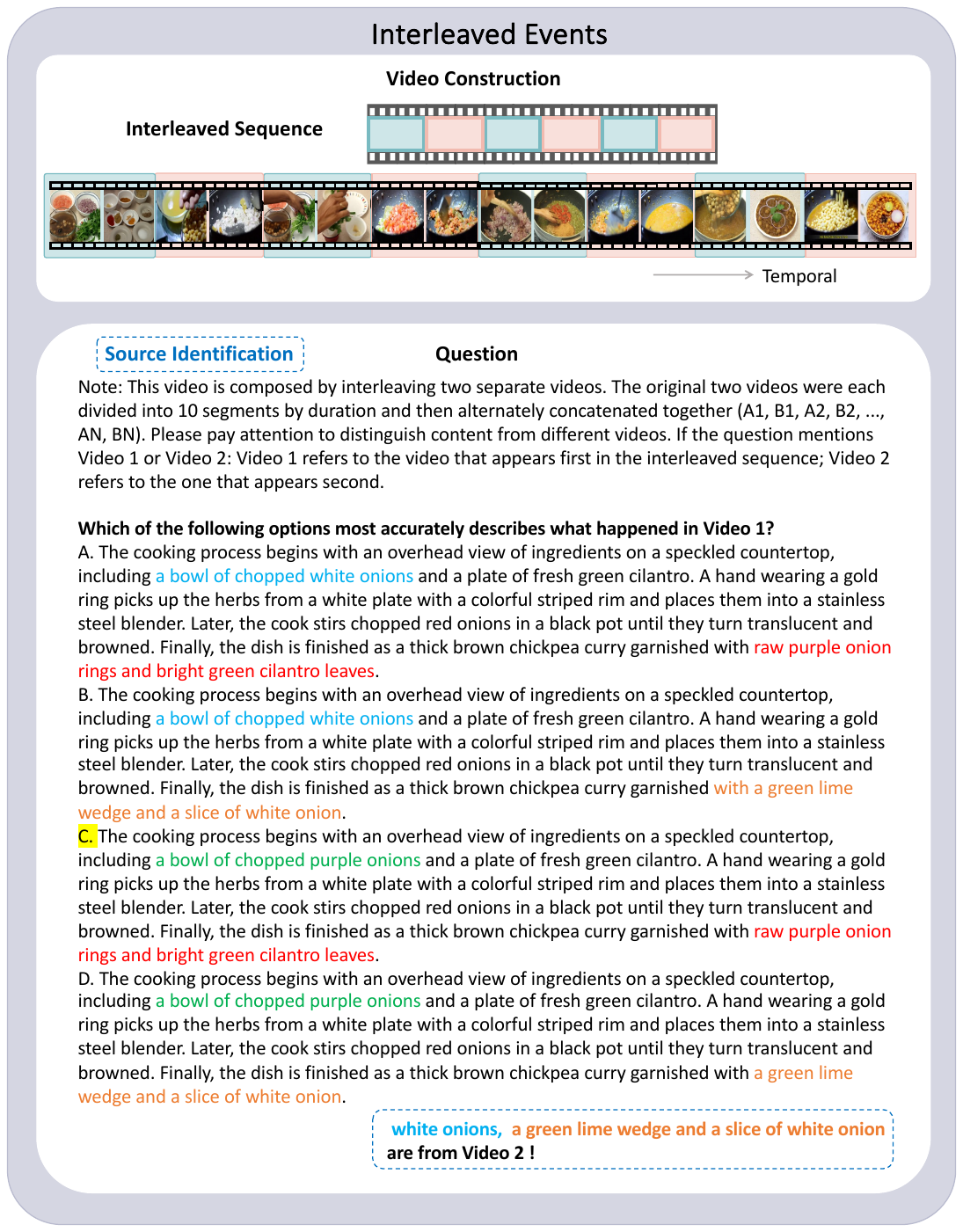}
    \vspace{-27mm}
    \caption{Example of Interleaved Events targeting \textit{Source Identification}. The three distractors replace certain content in the target video's narrative with content from the distractor video, while the correct option (highlighted in yellow) faithfully describes only the target video.}
    \label{fig:vis5}
    \vspace{-6mm}
\end{figure}
\begin{figure}[t]
    \centering
    \includegraphics[width=1\textwidth]{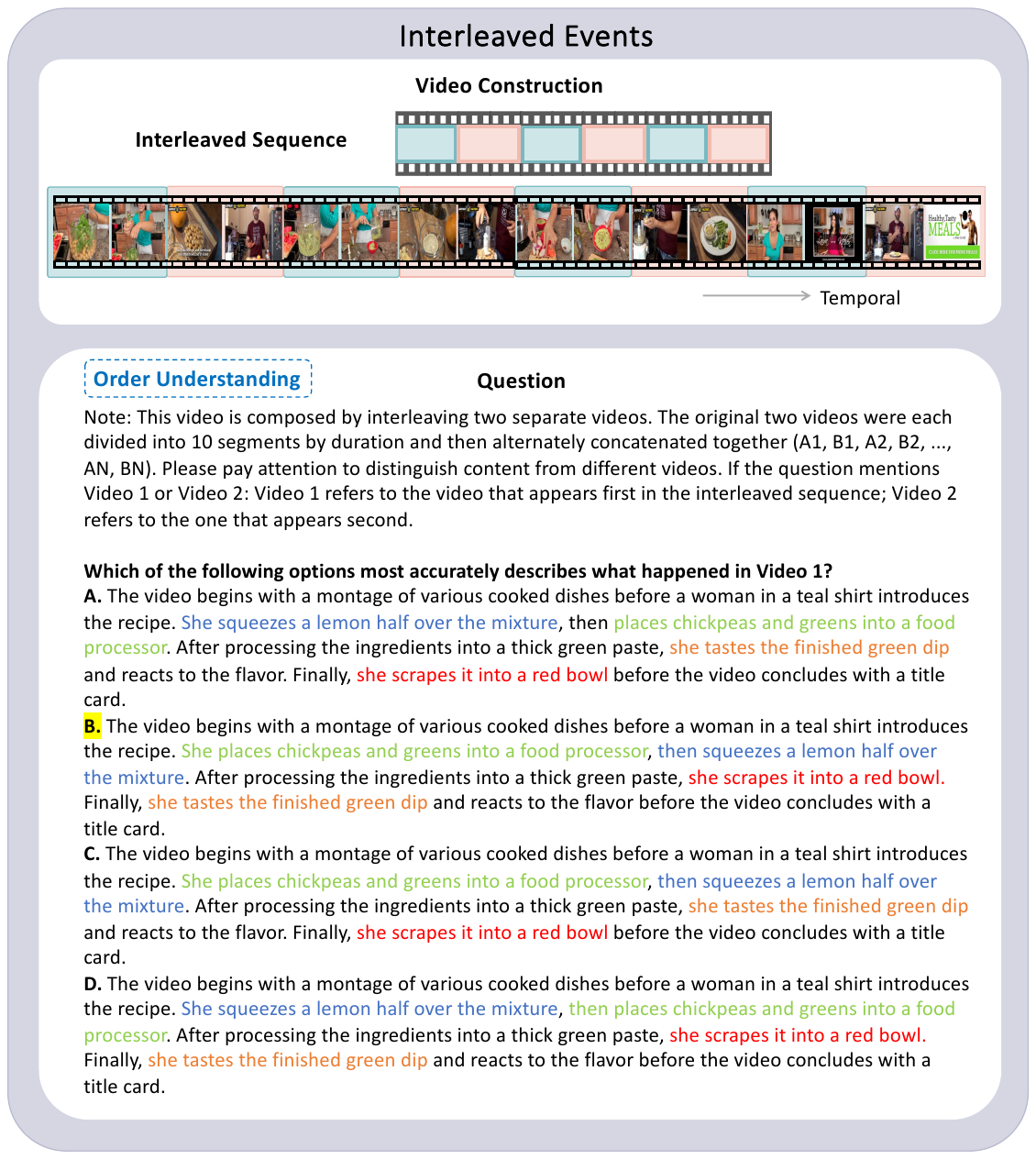}
    \vspace{-34mm}
    \caption{Example of Interleaved Events targeting \textit{Order Understanding}. The three distractors swap the temporal or logical sequence of events in the target video's narrative, while the correct option (highlighted in yellow) preserves the original order.}
    \label{fig:vis6}
    \vspace{-6mm}
\end{figure}
\begin{figure}[t]
    \centering
    \includegraphics[width=1\textwidth]{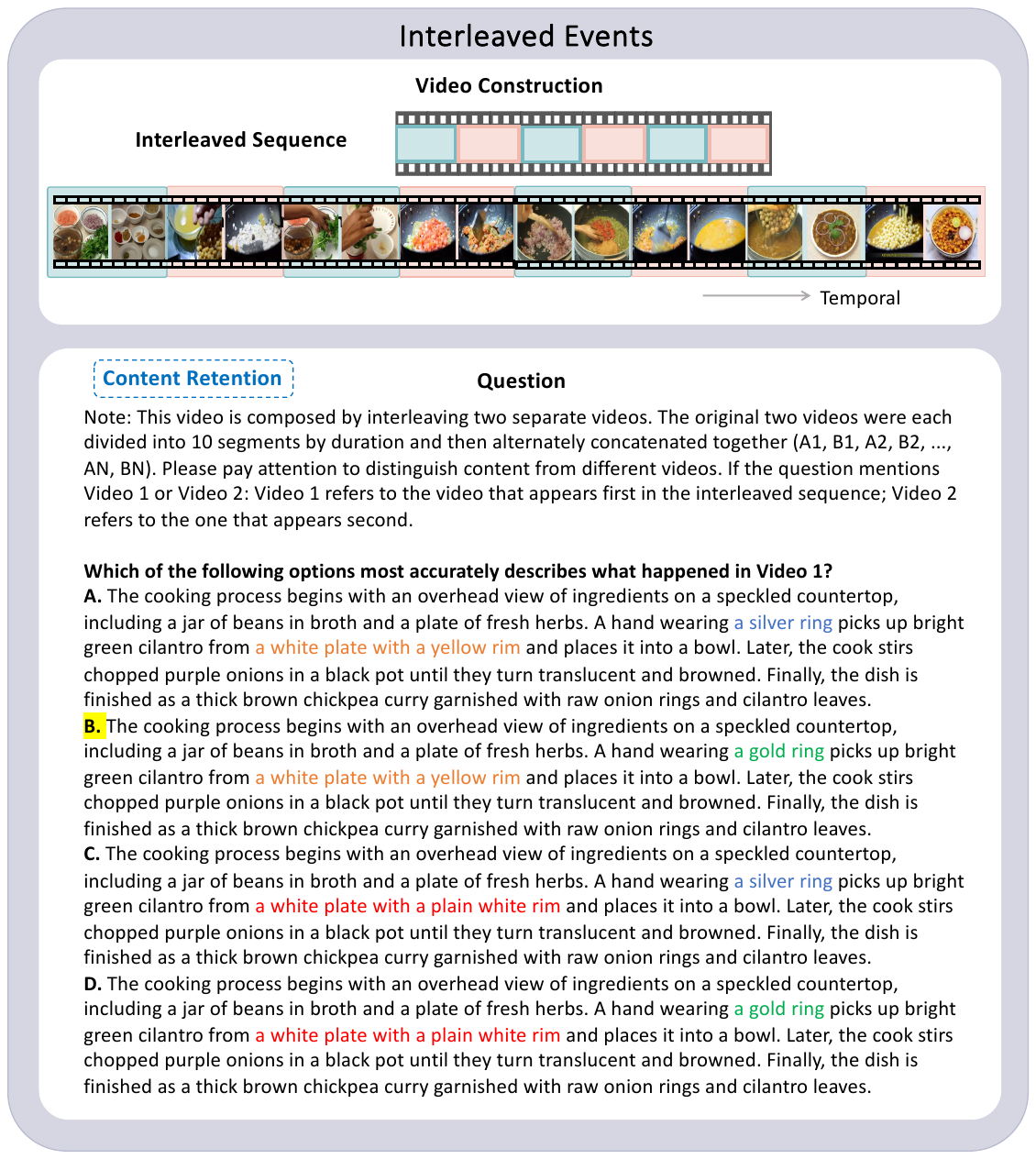}
    \vspace{-34mm}
    \caption{Example of Interleaved Events targeting \textit{Content Retention}. The three distractors replace certain content in the target video's narrative with plausible but fabricated content, while the correct option (highlighted in yellow) faithfully describes only the target video.}
    \label{fig:vis7}
    \vspace{-6mm}
\end{figure}
 
\begin{figure}[t]
    \centering
    \includegraphics[width=1\textwidth]{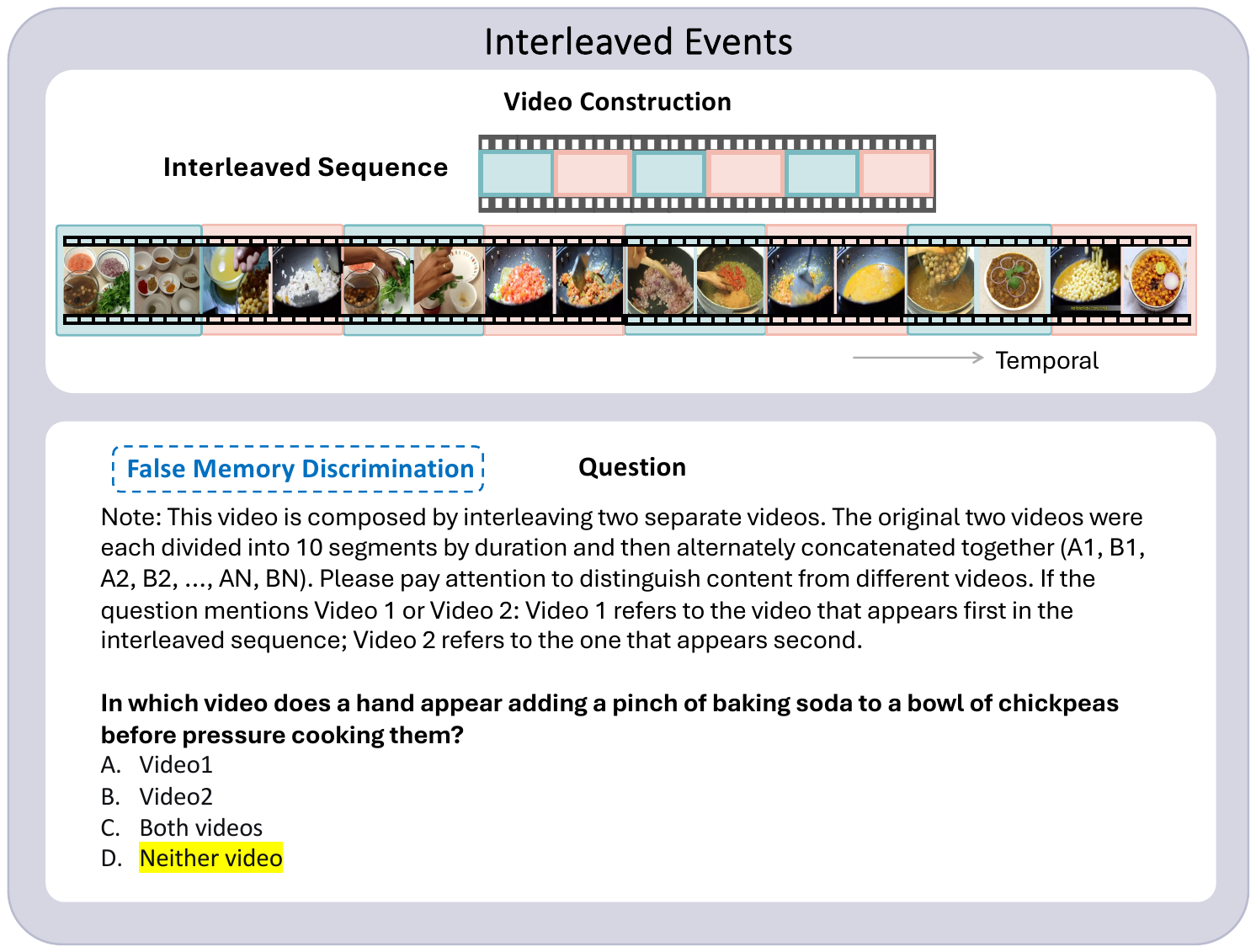}
    \vspace{-10mm}
    \caption{Example of Interleaved Events targeting \textit{False Memory Discrimination}. A fake question that is relevant to video content is presented, and the model should be aware to choose the option indicating that the query does not belong to either video. The correct answer is highlighted in yellow.}
    \label{fig:vis8}
    \vspace{-6mm}
\end{figure}

\begin{figure}[t]
    \centering
    \includegraphics[width=1\textwidth]{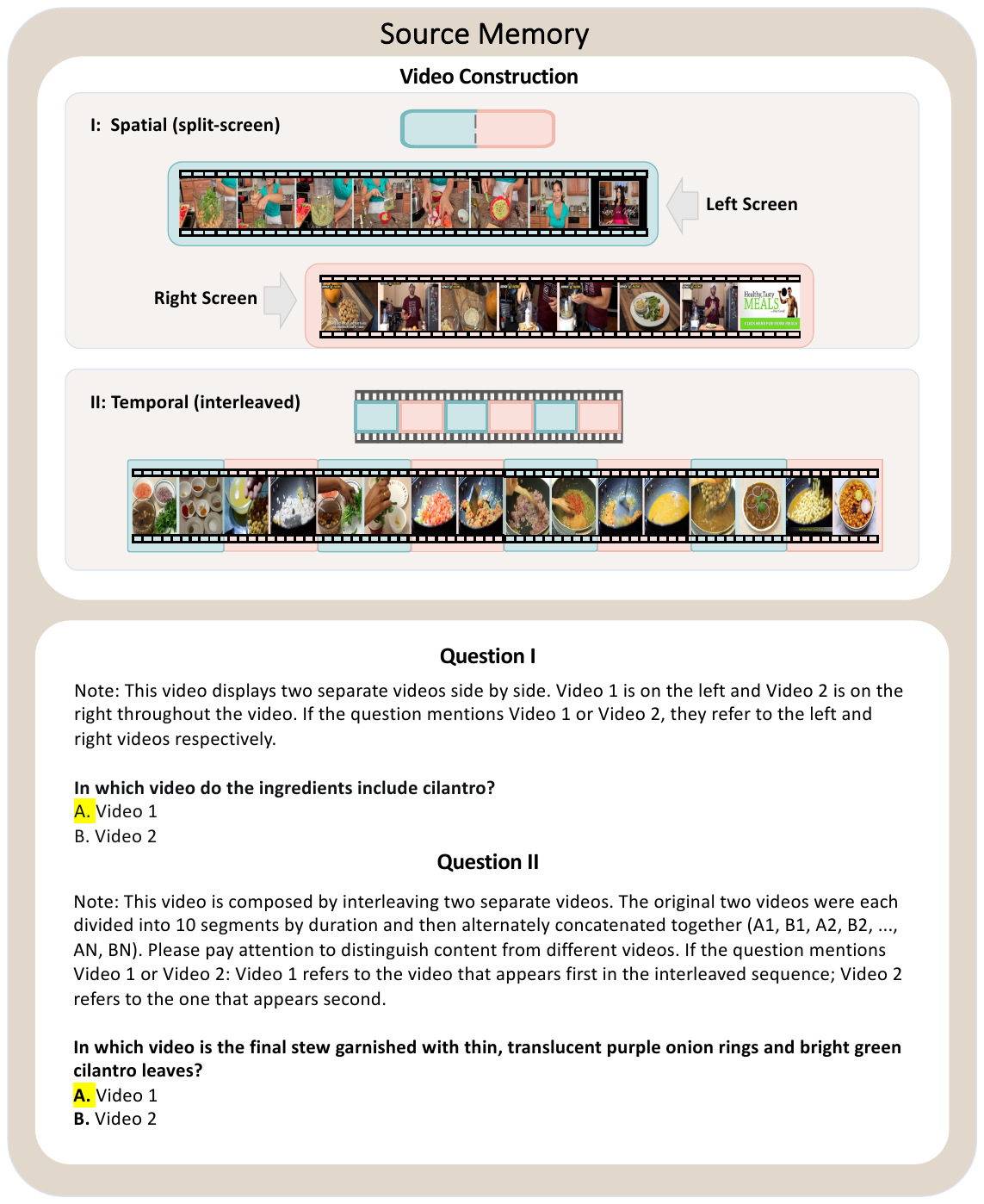}
    \vspace{-27mm}
    \caption{Example of Source Memory. Spatial refers to a split-screen format with frequent left/right swaps. The correct answer is highlighted in yellow.} 
    \label{fig:vis9}
    \vspace{-6mm}
\end{figure}

\begin{figure}[t]
    \centering
    \includegraphics[width=1\textwidth]{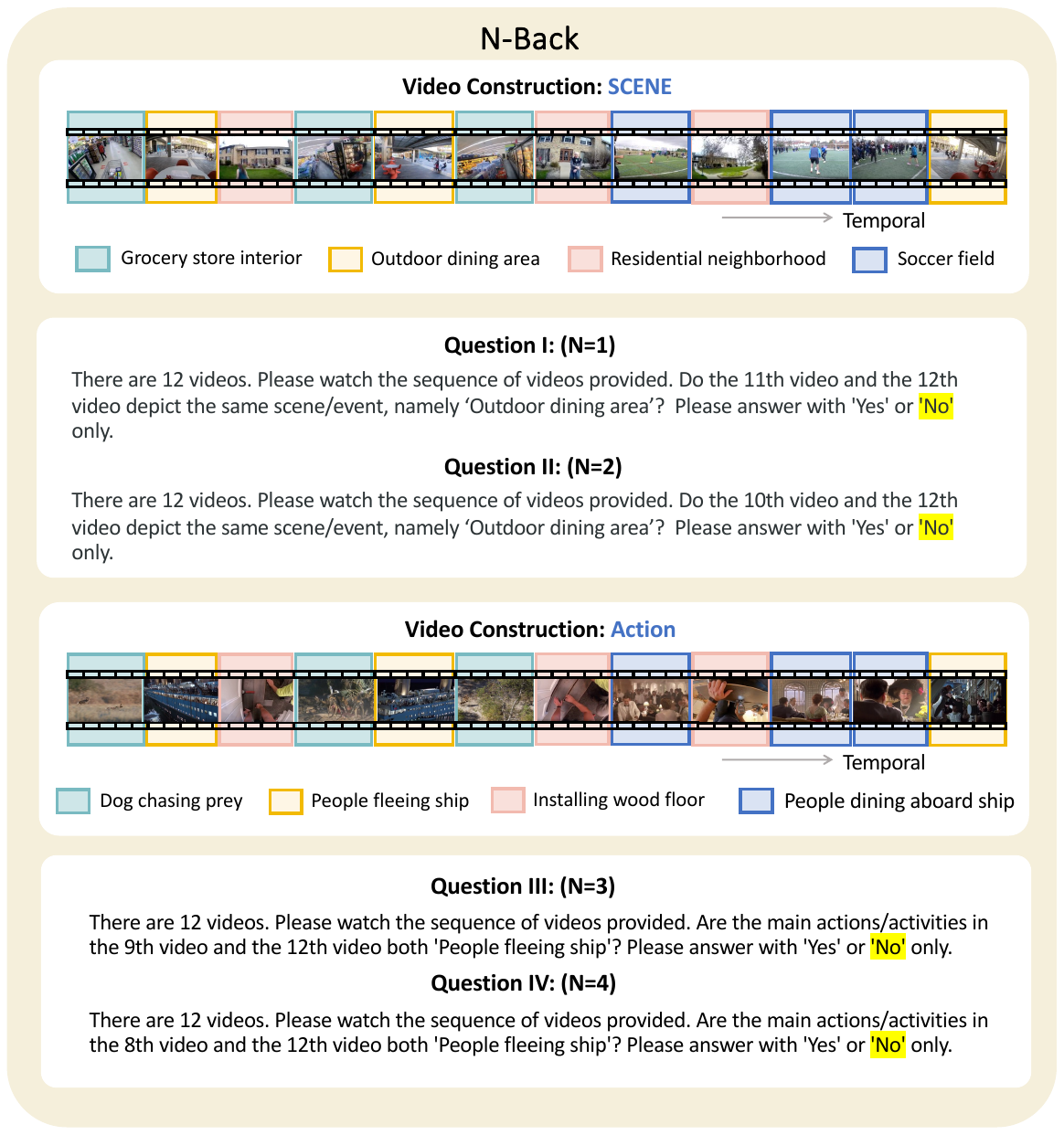}
    \vspace{-37mm}
    \caption{Example of N-Back. The model is asked to decide whether the final clip matches the clip $N$ positions earlier with a Yes/No answer, on two attributes: \textit{Scene} (same environment category) and \textit{Action} (same type of activity). The correct answer is highlighted in yellow.}
    \label{fig:vis10}
    \vspace{-6mm}
\end{figure}